# Making Decisions Using Sets of Probabilities: Updating, Time Consistency, and Calibration

**Peter D. Grünwald**                                                        PDG@CWI.NL
*CWI, P.O. Box 94079*
*1090 GB Amsterdam, The Netherlands*

**Joseph Y. Halpern**                                        HALPERN@CS.CORNELL.EDU
*Computer Science Department*
*Cornell University*
*Ithaca, NY 14853, USA*

## Abstract

We consider how an agent should update her beliefs when her beliefs are represented by a set $\mathcal{P}$ of probability distributions, given that the agent makes decisions using the *minimax criterion*, perhaps the best-studied and most commonly-used criterion in the literature. We adopt a game-theoretic framework, where the agent plays against a bookie, who chooses some distribution from $\mathcal{P}$. We consider two reasonable games that differ in what the bookie knows when he makes his choice. Anomalies that have been observed before, like *time inconsistency*, can be understood as arising because different games are being played, against bookies with different information. We characterize the important special cases in which the optimal decision rules according to the minimax criterion amount to either conditioning or simply ignoring the information. Finally, we consider the relationship between updating and *calibration* when uncertainty is described by sets of probabilities. Our results emphasize the key role of the *rectangularity condition* of Epstein and Schneider.

## 1. Introduction

Suppose that an agent models her uncertainty about a domain using a *set* $\mathcal{P}$ of probability distributions. How should the agent update $\mathcal{P}$ in light of observing that random variable $X$ takes on value $x$? Perhaps the standard answer is to condition each distribution in $\mathcal{P}$ on $X = x$ (more precisely, to condition those distributions in $\mathcal{P}$ that give $X = x$ positive probability on $X = x$), and adopt the resulting set of conditional distributions $\mathcal{P} \mid X = x$ as her representation of uncertainty. In contrast to the case where $\mathcal{P}$ is a singleton, it is often not clear whether conditioning is the right way to update a set $\mathcal{P}$. It turns out that in general, there is no single "right" way to update $\mathcal{P}$. Different updating methods satisfy different desiderata, and for some sets $\mathcal{P}$, not all of these desiderata can be satisfied at the same time. In this paper, we determine to what extent conditioning and some related update methods satisfy common decision-theoretic optimality properties. The main three questions we pose are:

1. Is conditioning the right thing to do under a *minimax* criterion, that is, does it lead to minimax-optimal decision rules?

2. Is the minimax criterion itself reasonable in the sense that it satisfies consistency criteria such as *time consistency* (defined formally below)?

3. Is conditioning the right thing do under a *calibration* criterion?

We show that the answer to the first two questions is "yes" if $\mathcal{P}$ satisfies a condition that Epstein and Schneider (2003) call *rectangularity*, while the answer to the third question is "yes" if $\mathcal{P}$ is convex and satisfies the rectangularity condition.[1] Thus, the main contribution of this paper is to show that, under the rectangularity condition, conditioning is the right thing to do under a wide variety of criteria. Apart from this main conclusion, our analysis provides new insights into the relation between minimax optimality, time consistency, and variants of conditioning (such as ignoring the information that $X = x$ altogether). We now discuss our contributions in more detail.

**1.1 The Minimax Criterion, Dilation, and Time Inconsistency**  How should an agent make decisions based on a set $\mathcal{P}$ of distributions?  Perhaps the best-studied and most commonly-used approach in the literature is to use the minimax criterion (Wald, 1950; Gärdenfors & Sahlin, 1982; Gilboa & Schmeidler, 1989). According to the minimax criterion, action $a_1$ is preferred to action $a_2$ if the worst-case expected loss of $a_1$ (with respect to all the probability distributions in the set $\mathcal{P}$ under consideration) is better than the worst-case expected loss of $a_2$. Thus, the action chosen is the one with the best worst-case outcome.

As has been pointed out by several authors, conditioning a set $\mathcal{P}$ on observation $X = x$ sometimes leads to a phenomenon called *dilation* (Augustin, 2003; Cozman & Walley, 2001; Herron, Seidenfeld, & Wasserman, 1997; Seidenfeld & Wasserman, 1993): the agent may have substantial knowledge about some other random variable $Y$ before observing $X = x$, but know significantly less after conditioning. Walley (1991, p. 299) gives a simple example of dilation: suppose that a fair coin is tossed twice, where the second toss may depend in an arbitrary way on the first.  (In particular, the tosses might be guaranteed to be identical, or guaranteed to be different.)  If $X$ represents the outcome of the first toss and $Y$ represents the outcome of the second toss, then before observing $X$, the agent believes that the probability that $Y$ is heads is $1/2$, while after observing $X$, the agent believes that the probability that $Y$ is heads can be an arbitrary element of $[0, 1]$.

While, as this example and others provided by Walley show, such dilation can be quite reasonable, it interacts rather badly with the minimax criterion, leading to anomalous behavior that has been called *time inconsistency* (Grünwald & Halpern, 2004; Seidenfeld, 2004): the minimax-optimal conditional decision rule before the value of $X$ is observed (which has the form "If $X = 0$ then do $a_1$; if $X = 1$ then do $a_2$; …") may be different from the minimax-optimal decision rule after conditioning.  For example, the minimax-optimal conditional decision rule may say "If $X = 0$ then do $a_1$", but the minimax-optimal decision rule conditional on observing $X = 0$ may be $a_2$.  (See Example 2.1.)  If uncertainty is modeled using a single distribution, such time inconsistency cannot arise.

**1.2 The Two Games**  To understand this phenomenon better, we model the decision problem as a game between the agent and a bookie (for a recent approach that is similar in spirit but done independently, see Ozdenoren & Peck, 2008). It turns out that there is more than one possible game that can be considered, depending on what information the bookie has. We focus on two (closely related) games here. In the first game, the bookie chooses a distribution from $\mathcal{P}$ before the agent moves. We show that the Nash equilibrium of this game leads to a minimax decision rule. (Indeed, this can be viewed as a justification of using the minimax criterion). However, in this game, conditioning on the information is not always optimal.[2] In the second game, the bookie gets to choose the distribution *after* the value of $X$ is observed. Again, in this game, the Nash equilibrium leads to the use of minimax, but now conditioning *is* the right thing to do.

---

1. All these results are proved under the assumption that the domain of the probability measures in $\mathcal{P}$ is finite and the set of actions that the decision maker is choosing among is finite.

If $\mathcal{P}$ is a singleton, the two games coincide (since there is only one choice the bookie can make, and the agent knows what it is). Not surprisingly, conditioning is the appropriate thing to do in this case. The moral of this analysis is that, when uncertainty is characterized by a set of distributions, if the agent is making decision using the minimax criterion, then the right decision depends on the game being played. The agent must consider if she is trying to protect herself against an adversary who knows the value of $X = x$ when choosing the distribution or one that does not know the value of $X = x$.

**1.3 Rectangularity and Time Consistency** In earlier work (Grünwald & Halpern, 2004) (GH from now on), we essentially considered the first game, and showed that, in this game, conditioning was not always the right thing to do when using the minimax criterion. Indeed, we showed there are sets $\mathcal{P}$ and games for which the minimax-optimal decision rule is to simply ignore the information. Our analysis of the first game lets us go beyond GH here in two ways. First, we provide a simple sufficient condition for when conditioning on the information is minimax optimal (Theorem 4.4). Second, we provide a sufficient condition for when it is minimax optimal to ignore information (Theorem 5.1).

Our sufficient condition guaranteeing that conditioning is minimax optimal can be viewed as providing a sufficient condition for time consistency. Our condition is essentially Epstein and Schneider's (2003) *rectangularity condition*, which they showed was sufficient to guarantee what has been called in the decision theory community *dynamic consistency*. Roughly speaking, dynamic consistency says that if, no matter what the agent learns, he will prefer decision rule $\delta$ to decision rule $\delta'$, then he should prefer $\delta$ to $\delta'$ before learning anything. Dynamic consistency is closely related to Savage's (1954) *sure-thing principle*. Epstein and Schneider show that, if an agent's uncertainty is represented using sets of probability distributions, all observations are possible (in our setting, this means that all probability distributions that the agent considers possible assign positive probability to all basic events of the form $X = x$), and the set of distributions satisfies the rectangularity condition then, no matter what the agent's loss function,[3] if the agent prefers $\delta$ to $\delta'$ after making an observation, then he will also prefer $\delta$ to $\delta'$ before making the observation. Conversely, they show that if the agent's preferences are dynamically consistent, then the agent's uncertainty can be represented by a set of probability measures that satisfies the rectangularity condition, and the agent can be viewed as making decisions using the minimax criterion.

Our results show that if all observations are possible and the rectangularity condition holds, then, no matter what the loss function, *time consistency* holds. Time consistency holds if a decision is minimax optimal before making an observation iff it is optimal after making the observation. Note that time consistency just considers just the optimal decision, while dynamic consistency considers the whole preference order. However, time consistency is an "iff" requirement: a decision is optimal before making the observation *if and only if* that decision is optimal after making the observation. By way of contrast, dynamic consistency is uni-directional: if $a$ is preferred to $a'$ after making the observation, then it must still be preferred before making the observation.

---

2. In some other senses of the words "conditioning" and "optimal," conditioning on the information *is* always optimal. This is discussed further in Section 7.

3. We work with loss functions in this paper rather than utility functions, since losses seem to be somewhat more standard in this literature. However, we could trivially restate our results in terms of utility.

These results show that if uncertainty is represented by a rectangular set of measures, all observations are possible, and the minimax criterion is used, then both dynamic consistency and time consistency hold. On the other hand, as we show in Proposition 4.7, in general dynamic consistency and time consistency are incomparable.

**1.4 $\mathcal{C}$-conditioning and Calibration** As stated, we provide a general condition on $\mathcal{P}$ under which conditioning is minimax optimal, as well as a general condition under which ignoring the information is minimax optimal. Note that ignoring the information can also be viewed as the result of conditioning; not conditioning on the information, but conditioning on the whole space. This leads us to consider a generalization of conditioning. Let $\mathcal{C}$ be a partition of the set of values of the random variable $X$, and let $\mathcal{C}(x)$ be the element of the partition that contains $x$. Suppose that when we observe $x$, we condition on the event $X \in \mathcal{C}(x)$. We call this variant of conditioning $\mathcal{C}$-conditioning; standard conditioning is just the special case where each element of $\mathcal{C}$ is a singleton. Is $\mathcal{C}$-conditioning always minimax optimal in the first game? That is, is it always optimal to condition on *something*? As we show by considering a variation of the Monty Hall Problem (Example 5.4), this is not the case in general.

Nevertheless, it turns out that considering $\mathcal{C}$-conditioning is useful; it underlies our analysis of *calibration*. As pointed out by Dawid (1982), an agent updating her beliefs and making decisions on the basis of these beliefs should also be concerned about being calibrated. Calibration is usually defined in terms of empirical data. To explain what it means and its connection to decision making, consider an agent that is a weather forecaster on your local television station. Every night the forecaster makes a prediction about whether or not it will rain the next day in the area where you live. She does this by asserting that the probability of rain is $p$, where $p \in \{0, 0.1, \ldots, 0.9, 1\}$. How should we interpret these probabilities? The usual interpretation is that, in the long run, on those days at which the weather forecaster predict probability $p$, it will rain approximately $100p\%$ of the time. Thus, for example, among all days for which she predicted 0.1, the fraction of days with rain was close to 0.1. A weather forecaster with this property is said to be *calibrated*. If a weather forecaster is calibrated, and you make bets which, based on her probabilistic predictions, seem favorable, then in the long run you cannot lose money. On the other hand, if a weather forecaster is not calibrated, there exist bets that may seem favorable but result in a loss. So clearly there is a close connection between calibration and decision making.

Calibration is usually defined relative to empirical data or singleton distributions. We first consider the obvious extension to sets of probabilities, but the obvious extension turns out to be only a very weak requirement. We therefore define a stronger and arguably more interesting variation that we call *sharp* calibration. We take an *update rule* $\Pi$ to map a set $\mathcal{P}$ and a value $x$ to a new set $\Pi(\mathcal{P}, x)$ of probabilities. Intuitively, $\Pi(\mathcal{P}, x)$ is the result of updating $\mathcal{P}$ given the observation $X = x$, according to update rule $\Pi$. A calibrated update rule $\Pi$ is sharply calibrated for $\mathcal{P}$ if there is no other rule $\Pi'$ that is also calibrated such that, for all $x$, $\Pi'(\mathcal{P}, x) \subset \Pi(\mathcal{P}, x)$, and for some $x$, the inclusion is strict. We first show that if $\mathcal{P}$ is convex, then $\mathcal{C}$-conditioning is sharply calibrated for some $\mathcal{C}$; different choices of $\mathcal{P}$ require different $\mathcal{C}$. We then show that, if $\mathcal{P}$ also satisfies the rectangularity condition, then standard conditioning is sharply calibrated.

**1.5 Discussion**   Both the idea of representing uncertainty by a set $\mathcal{P}$ of distributions and that of handling decisions in a worst-case optimal manner may, of course, be criticized. While we do not claim that this is necessarily the "right" or the "best" approach, it is worth pointing out that two of the most common criticisms are, to some extent, unjustified. First, since it may be hard for an agent to determine the precise boundaries of the set $\mathcal{P}$, it has been argued that "soft boundaries" are more appropriate.   These soft boundaries may be thought of as inducing a single distribution on $\Delta(\mathcal{X} \times \mathcal{Y})$, the set of probability distributions on $\mathcal{X} \times \mathcal{Y}$ (with the density of $\Pr \in \Delta(\mathcal{X} \times \mathcal{Y})$ proportional to the extent to which "$\Pr$ is included in the set $\mathcal{P}$"). With this single distribution, the setting becomes equivalent to the setting of standard Bayesian decision theory. The problem with this criticism is that in some cases, hard boundaries are in fact natural. For example, some conditional probabilities may be known to be precisely 0, as is the case in the Monty Hall game (Example 5.4). Similarly, the use of the minimax criterion is not as pessimistic as is often thought. The minimax solution often coincides with the Bayes-optimal solution under some "maximum entropy" prior (Grünwald & Dawid, 2004), which is not commonly associated with being overly pessimistic. In fact, in the Monty Hall problem, the minimax-optimal decision rule coincides with the solution usually advocated, which requires making further assumptions about $\mathcal{P}$ to reduce it to a singleton.

The rest of this paper is organized as follows. In Section 2, we define the basic framework. In Section 3, we formally define the two games described above and show that the minimax-optimal decision rule gives a Nash equilibrium. In Section 4, we characterize the minimax-optimal decision rule for the first game, in which the bookie chooses a distribution before $X$ is observed. In Section 5 we discuss $\mathcal{C}$-conditioning and show that, in general, it is not minimax optimal. In Section 6, we discuss calibration and $\mathcal{C}$-conditioning. We conclude with some discussion in Section 7. All proofs can be found in the appendix.

## 2. Notation and Definitions

In this paper, uncertainty is represented by a set $\mathcal{P}$ of probability distributions. For ease of exposition, we assume throughout this paper that we are interested in two random variables, $X$ and $Y$, which can take values in spaces $\mathcal{X}$ and $\mathcal{Y}$, respectively. $\mathcal{P}$ always denotes a set of distributions on $\mathcal{X} \times \mathcal{Y}$; that is, $\mathcal{P} \subseteq \Delta(\mathcal{X} \times \mathcal{Y})$, where $\Delta(S)$ denotes the set of probability distributions on $S$. For ease of exposition, we assume that $\mathcal{P}$ is a closed set; this is a standard assumption in the literature that seems quite natural in our applications, and makes the statement of our results simpler (otherwise we have to state our results using closures). If $\Pr \in \Delta(\mathcal{X} \times \mathcal{Y})$, let $\Pr_{\mathcal{X}}$ and $\Pr_{\mathcal{Y}}$ denote the marginals of $\Pr$ on $\mathcal{X}$ and $\mathcal{Y}$, respectively. Let $\mathcal{P}_{\mathcal{Y}} = \{\Pr_{\mathcal{Y}} : \Pr \in \mathcal{P}\}$. If $E \subseteq \mathcal{X} \times \mathcal{Y}$, then let $\mathcal{P} \mid E = \{\Pr \mid E : \Pr \in \mathcal{P}, \Pr(E) > 0\}$. Here

Pr | $E$ (often written Pr($\cdot$ | $E$)) is the distribution on $\mathcal{X} \times \mathcal{Y}$ obtained by conditioning on $E$.

The represesenatation of uncertainty using sets of probability distributions is closely related to Walley's (1991) use of *(lower and upper) previsions*. A prevision is an expectation function; that is, a lower prevision is a mapping random variables to the reals satisfying certain properties. It is well known (Huber, 1981) that what Walley calls a *coherent* lower prevision (a lower prevision satisfying some minimal properties) can be identified with the *lower expectation* of a set of probability measures (that is, the function $E$ such that $E(X) = \inf_{\mathrm{Pr} \in \mathcal{P}} E_{\mathrm{Pr}}(X)$). Indeed, there is a one-to-one map between lower previsions and closed convex sets of probability measures. The notion of conditioning we are using corresponds to what Walley calls the *regular extension* of a lower prevision (see Walley, 1991, Appendix J).

**2.1 Loss Functions**  As in GH, we are interested in an agent who must choose some action from a set $\mathcal{A}$, where the loss of the action depends only on the value of random variable $Y$. We assume in this paper that $\mathcal{X}$, $\mathcal{Y}$, and $\mathcal{A}$ are finite, and that $|\mathcal{A}| \geq 2$, so that there are always at least two possible choices. (If we allowed $\mathcal{A}$ to be a singleton, then some of our results would not hold for trivial reasons.)

We assume that with each action $a \in \mathcal{A}$ and value $y \in \mathcal{Y}$ is associated some loss to the agent. (The losses can be negative, which amounts to a gain.) Let $L : \mathcal{Y} \times \mathcal{A} \to I\!\!R$ be the loss function.

Such loss functions arise quite naturally. For example, in a medical setting, we can take $\mathcal{Y}$ to consist of the possible diseases and $\mathcal{X}$ to consist of symptoms. The set $\mathcal{A}$ consists of possible courses of treatment that a doctor can choose. The doctor's loss function depends only on the patient's disease and the course of treatment, not on the symptoms. But, in general, the doctor's choice of treatment depends on the symptoms observed.

**2.3 Decision Problems and Decision Settings**  For our purposes, a *decision setting* is a tuple $DS = (\mathcal{X}, \mathcal{Y}, \mathcal{A}, \mathcal{P})$, where $\mathcal{X}$, $\mathcal{Y}$, $\mathcal{A}$, and $\mathcal{P}$ are as above. A *decision problem* is characterized by a tuple $DP = (\mathcal{X}, \mathcal{Y}, \mathcal{A}, \mathcal{P}, L)$, where $L$ is a loss function. That is, a decision problem is a decision setting together with a loss function. We say that the decision problem $(\mathcal{X}, \mathcal{Y}, \mathcal{A}, \mathcal{P}, L)$ is *based on* the decision setting $(\mathcal{X}, \mathcal{Y}, \mathcal{A}, \mathcal{P})$.

**2.4 Decision Rules**  Given a decision problem $DP = (\mathcal{X}, \mathcal{Y}, \mathcal{A}, \mathcal{P}, L)$, suppose that the agent observes the value of the variable $X$. After having observed $X$, she must perform an act, the quality of which is judged according to loss function $L$. The agent must choose a *decision rule* that determines what she does as a function of her observations. We allow decision rules to be randomized. Thus, a decision rule is a function $\delta : \mathcal{X} \to \Delta(\mathcal{A})$ that chooses a distribution over actions based on the agent's observations. Let $\mathcal{D}(\mathcal{X}, \mathcal{A})$ be the set of all decision rules. A special case is a deterministic decision rule, which assigns probability 1 to a particular action. If $\delta$ is deterministic, we sometimes abuse notation and write $\delta(x)$ for the action that is assigned probability 1 by the distribution $\delta(x)$. Given a decision rule $\delta$ and a loss function $L$, let $L_\delta$ be the random variable on $\mathcal{X} \times \mathcal{Y}$ such that $L_\delta(x, y) = \sum_{a \in \mathcal{A}} \delta(x)(a) L(y, a)$. Here $\delta(x)(a)$ stands for the probability of performing action $a$ according to the distribution $\delta(x)$ over actions that is adopted when $x$ is observed. Note that in the special case that $\delta$ is a deterministic decision rule, $L_\delta(x, y) = L(y, \delta(x))$.

We also extend this notation to randomized actions: for $\alpha \in \Delta(\mathcal{A})$, we let $L_\alpha$ be the random variable on $\mathcal{Y}$ such that $L_\alpha(y) = \sum_{a \in \mathcal{A}} \alpha(a) L(y, a)$.

A decision rule $\delta^0$ is *a priori minimax optimal* for the decision problem $DP$ if

$$\max_{\Pr \in \mathcal{P}} E_{\Pr}[L_{\delta^0}] = \min_{\delta \in \mathcal{D}(\mathcal{X}, \mathcal{A})} \max_{\Pr \in \mathcal{P}} E_{\Pr}[L_\delta].$$

That is, $\delta^0$ is a priori minimax optimal if $\delta^0$ gives the best worst-case expected loss with respect to all the distributions in Pr. We can write max here instead of sup because of our assumption that $\mathcal{P}$ is closed. This ensures that there is some $\Pr \in \mathcal{P}$ for which $E_{\Pr}[L_{\delta^0}]$ takes on its maximum value.

A decision rule $\delta^1$ is *a posteriori minimax optimal* for $DP$ if, for all $x \in \mathcal{X}$ such that $\Pr(X = x) > 0$ for some $\Pr \in \mathcal{P}$,

$$\max_{\Pr \in \mathcal{P} | X = x} E_{\Pr}[L_{\delta^1}] = \min_{\delta \in \mathcal{D}(\mathcal{X}, \mathcal{A})} \max_{\Pr \in \mathcal{P} | X = x} E_{\Pr}[L_\delta]. \tag{1}$$

To get the a posteriori minimax-optimal decision rule we do the obvious thing: if $x$ is observed, we simply condition each probability distribution $\Pr \in \mathcal{P}$ on $X = x$, and choose the action that gives the least expected loss (in the worst case) with respect to $\mathcal{P} \mid X = x$. Since all distributions Pr mentioned in (1) satisfy $\Pr(X = x) = 1$, the minimum over $\delta \in \mathcal{D}(\mathcal{X}, \mathcal{A})$ does not depend on the values of $\delta(x')$ for $x' \neq x$; the minimum is effectively over randomized actions rather than decision rules.

As the following example, taken from GH, shows, a priori minimax-optimal decision rules are in general different from a posteriori minimax-optimal decision rules.

**Example 2.1:** Suppose that $\mathcal{X} = \mathcal{Y} = \mathcal{A} = \{0, 1\}$ and $\mathcal{P} = \{\Pr \in \Delta(\mathcal{X} \times \mathcal{Y}) : \Pr_{\mathcal{Y}}(Y = 1) = 2/3\}$. Thus, $\mathcal{P}$ consists of all distributions whose marginal on $Y$ gives $Y = 1$ probability $2/3$. We can think of the actions in $\mathcal{A}$ as predictions of the value of $Y$. The loss function is 0 if the right value is predicted and 1 otherwise; that is, $L(i, j) = |i - j|$. This is the so-called $0/1$ or *classification* loss. It is easy to see that the optimal a priori decision rule is to choose 1 no matter what is observed (which has expected loss $1/3$). Intuitively, observing the value of $X$ tells us nothing about the value of $Y$, so the best decision is to predict according to the prior probability of $Y = 1$. However, all probabilities on $Y = 1$ are compatible with observing either $X = 0$ or $X = 1$. That is, both $(\mathcal{P} \mid X = 0)_{\mathcal{Y}}$ and $(\mathcal{P} \mid X = 1)_{\mathcal{Y}}$ consist of all distributions on $\mathcal{Y}$. Thus, the minimax optimal a posteriori decision rule randomizes (with equal probability) between $Y = 0$ and $Y = 1$.

To summarize, if you make decisions according to the minimax optimality criterion, then before making an observation, you will predict $Y = 1$. However, *no matter what observation you make*, after making the observation, you will randomize (with equal probability) between predicting $Y = 0$ and $Y = 1$. Moreover, you know even before making the observation that your opinion as to the best decision rule will change in this way. (Note that this is an example of both time inconsistency and dynamic inconsistency.) ∎

**2.5 Time and Dynamic Consistency**    Formally, a decision problem $DP$ is *time consistent* iff, for all decision rules $\delta$, $\delta$ is a priori minimax optimal for $DP$ iff $\delta$ is a posteriori minimax optimal. We say that $DP$ is *weakly time consistent* if every a posteriori minimax optimal rule for $DP$ is also a priori minimax optimal for $DP$. A decision setting $DS$ is (weakly) time consistent if every decision problem based on $DS$ is.

Following Epstein and Schneider (2003), we say that a decision problem $DP$ is *dynamically consistent* if for every pair $\delta, \delta'$ of decision rules, the following conditions both hold:

1. If, for all $x$ such that $\Pr(X = x) > 0$ for some $\Pr \in \mathcal{P}$,

$$\max_{\Pr \in (\mathcal{P}|X=x)} E_{\Pr}[L_\delta] \leq \max_{\Pr \in (\mathcal{P}|X=x)} E_{\Pr}[L_{\delta'}], \qquad (2)$$

then

$$\max_{\Pr \in \mathcal{P}} E_{\Pr}[L_\delta] \leq \max_{\Pr \in \mathcal{P}} E_{\Pr}[L_{\delta'}]. \qquad (3)$$

2. If, for all $x$ such that $\Pr(X = x) > 0$ for some $\Pr \in \mathcal{P}$, we have strict inequality in (2), then (3) must hold with strict inequality as well.

Informally, dynamic consistency means that whenever $\delta$ is preferred to $\delta'$ according to the minimax criterion a posteriori, then $\delta$ is also preferred to $\delta'$ according to the minimax criterion a priori, and that whenever the a posteriori preference is strict for all possible observations, then the a priori preference must be strict as well.

A decision setting $DS$ is dynamically consistent if every decision problem based on $DS$ is.

## 3. Two Game-Theoretic Interpretations of $\mathcal{P}$

What does it mean that an agent's uncertainty is characterized by a set $\mathcal{P}$ of probability distributions? How should we understand $\mathcal{P}$? We give $\mathcal{P}$ a game-theoretic interpretation here: namely, an adversary gets to choose a distribution from the set $\mathcal{P}$.[4] But this does not completely specify the game. We must also specify *when* the adversary makes the choice. We consider two times that the adversary can choose: the first is before the agents observes the value of $\mathcal{X}$, and the second is after. We formalize this as two different games, where we take the "adversary" to be a bookie.

We call the first game the $\mathcal{P}$-game. It is defined as follows:

1. The bookie chooses a distribution $\Pr \in \mathcal{P}$.
2. The value $x$ of $X$ is chosen (by nature) according to $\Pr_{\mathcal{X}}$ and observed by both bookie and agent.
3. The agent chooses an action $a \in \mathcal{A}$.
4. The value $y$ of $Y$ is chosen according to $\Pr \mid X = x$.
5. The agent's loss is $L(y, a)$; the bookie's loss is $-L(y, a)$.

This is a zero-sum game; the agent's loss is the bookie's gain. In this game, the agent's strategy is a decision rule, that is, a function that gives a distribution over actions for each observed value of $X$. The bookie's strategy is a distribution over distributions in $\mathcal{P}$.

We now consider a second interpretation of $\mathcal{P}$, characterized by a different game that gives the bookie more power. Rather than choosing the distribution before observing the value of $X$, the bookie gets to choose the distribution after observing the value. We call this the $\mathcal{P}$-$X$-game. Formally, it is specified as follows:

---

4. This interpretation remains meaningful in several practical situations where there is no explicit adversary; see the final paragraph of this section.

1. The value $x$ of $X$ is chosen (by nature) according to some procedure that is guaranteed to end up with a value of $x$ for which $\Pr(X = x) > 0$ for some $\Pr \in \mathcal{P}$, and observed by both the bookie and the agent.[5]

2. The bookie chooses a distribution $\Pr \in \mathcal{P}$ such that $\Pr(X = x) > 0$.[6]

3. The agent chooses an action $a \in \mathcal{A}$.

4. The value $y$ of $Y$ is chosen according to $\Pr \mid X = x$.

5. The agent's loss is $L(y, a)$; the bookie's loss is $-L(y, a)$.

Recall that a pair of strategies $(S_1, S_2)$ is a Nash equilibrium if neither party can do better by unilaterally changing strategies. If, as in our case, $(S_1, S_2)$ is a Nash equilibrium in a zero-sum game, it is also known as a "saddle point"; $S_1$ must be a minimax strategy, and $S_2$ must be a maximin strategy (Grünwald & Dawid, 2004). As the following results show, an agent must be using an a priori minimax-optimal decision rule in a Nash equilibrium of the $\mathcal{P}$-game, and an a posteriori minimax-optimal decision rule is a Nash equilibrium of the $\mathcal{P}$-$X$-game. This can be viewed as a justification for using (a priori and a posteriori) minimax-optimal decision rules.

**Theorem 3.1:** *Fix $\mathcal{X}$, $\mathcal{Y}$, $\mathcal{A}$, $L$, and $\mathcal{P} \subseteq \Delta(\mathcal{X} \times \mathcal{Y})$.*

(a) *The $\mathcal{P}$-game has a Nash equilibrium $(\pi^*, \delta^*)$, where $\pi^*$ is a distribution over $\mathcal{P}$ with finite support.*

(b) *If $(\pi^*, \delta^*)$ is a Nash equilibrium of the $\mathcal{P}$-game such that $\pi^*$ has finite support, then*

    (i) *for every distribution $\Pr' \in \mathcal{P}$ in the support of $\pi^*$, we have*
$E_{\Pr'}[L_{\delta^*}] = \max_{\Pr \in \mathcal{P}} E_{\Pr}[L_{\delta^*}];$

    (ii) *if $\Pr^* = \sum_{\Pr \in \mathcal{P}, \pi^*(\Pr) > 0} \pi^*(\Pr) \Pr$ (i.e., $\Pr^*$ is the convex combination of the distributions in the support of $\pi^*$, weighted by their probability according to $\pi^*$), then*

$$
\begin{aligned}
E_{\Pr^*}[L_{\delta^*}] &= \min_{\delta \in \mathcal{D}(\mathcal{X}, \mathcal{A})} E_{\Pr^*}[L_\delta] \\
&= \max_{\Pr \in \mathcal{P}} \min_{\delta \in \mathcal{D}(\mathcal{X}, \mathcal{A})} E_{\Pr}[L_\delta] \\
&= \min_{\delta \in \mathcal{D}(\mathcal{X}, \mathcal{A})} \max_{\Pr \in \mathcal{P}} E_{\Pr}[L_\delta] \\
&= \max_{\Pr \in \mathcal{P}} E_{\Pr}[L_{\delta^*}].
\end{aligned}
$$

Once nature has chosen a value for $X$ in the $\mathcal{P}$-$X$-game, we can regard steps 2–5 of the $\mathcal{P}$-$X$-game as a game between the bookie and the agent, where the bookie's strategy is characterized by a distribution in $\mathcal{P} \mid X = x$ and the agent's is characterized by a distribution over actions. We call this the $\mathcal{P}$-$x$-game.

**Theorem 3.2:** *Fix $\mathcal{X}$, $\mathcal{Y}$, $\mathcal{A}$, $L$, $\mathcal{P} \subseteq \Delta(\mathcal{X} \times \mathcal{Y})$.*

---

5. Because $x$ is observed by both parties, and $y$ is chosen after $x$ is chosen, the procedure by which nature chooses $x$ is irrelevant. We could assume for definiteness that nature chooses uniformly at random among the values $x$ such that $\Pr(x) > 0$ for some $\Pr \in \mathcal{P}$, but any other choice would work equally well.

6. If we were to consider *conditional probability distributions* (de Finetti, 1936; Popper, 1968), for which $\Pr(Y = y \mid X = x)$ is defined even if $\Pr(X = x) = 0$, then we could drop the restriction that $x$ is chosen such that $\Pr(X = x) > 0$ for some $\Pr \in \mathcal{P}$.

(a) The $\mathcal{P}$-$x$-game has a Nash equilibrium $(\pi^*, \delta^*(x))$, where $\pi^*$ is a distribution over $\mathcal{P} \mid X = x$ with finite support.

(b) If $(\pi^*, \delta^*(x))$ is a Nash equilibrium of the $\mathcal{P}$-$x$-game such that $\pi^*$ has finite support, then

  (i) for all $\mathrm{Pr}'$ in the support of $\pi^*$, we have $E_{\mathrm{Pr}'}[L_{\delta^*}] = \max_{\mathrm{Pr} \in \mathcal{P} \mid X = x} E_{\mathrm{Pr}}[L_{\delta^*}]$;

  (ii) if $\mathrm{Pr}^* = \sum_{\mathrm{Pr} \in \mathcal{P}, \pi^*(\mathrm{Pr}) > 0} \pi^*(\mathrm{Pr}) \, \mathrm{Pr}$, then

$$
\begin{aligned}
E_{\mathrm{Pr}^*}[L_{\delta^*}] &= \min_{\delta \in \mathcal{D}(\mathcal{X}, \mathcal{A})} E_{\mathrm{Pr}^*}[L_\delta] \\
&= \max_{\mathrm{Pr} \in \mathcal{P} \mid X = x} \min_{\delta \in \mathcal{D}(\mathcal{X}, \mathcal{A})} E_{\mathrm{Pr}}[L_\delta] \\
&= \min_{\delta \in \mathcal{D}(\mathcal{X}, \mathcal{A})} \max_{\mathrm{Pr} \in \mathcal{P} \mid X = x} E_{\mathrm{Pr}}[L_\delta] \\
&= \max_{\mathrm{Pr} \in \mathcal{P} \mid X = x} E_{\mathrm{Pr}}[L_{\delta^*}].
\end{aligned}
$$

Since all distributions $\mathrm{Pr}$ in the expression $\min_{\delta \in \mathcal{D}(\mathcal{X}, \mathcal{A})} \max_{\mathrm{Pr} \in \mathcal{P} \mid X = x} E_{\mathrm{Pr}}[L_\delta]$ in part (b)(ii) are in $\mathcal{P} \mid X = x$, as in (1), the minimum is effectively over randomized actions rather than decision rules.

Theorems 3.1 and 3.2 can be viewed as although, according to the definition, there is time inconsistency, when viewed properly, there is no real inconsistency here; rather, we must just be careful about what game is being played. If the $\mathcal{P}$-game is being played, the right strategy is the a priori minimax-optimal strategy, both before and after the value of $X$ is observed; similarly, if the $\mathcal{P}$-$X$-game is being played, the right strategy is the a posteriori minimax-optimal strategy, both before and after the value of $X$ is observed. Indeed, thinking in terms of the games explains the apparent time inconsistency. In both games, the agent gains information by observing $X = x$. But in the $\mathcal{P}$-$X$ game, so does the bookie. The information may be of more use to the bookie than the agent, so, in this game, the agent can be worse off by being given the opportunity to learn the value of $X$.

Of course, in most practical situations, agents (robots, statisticians, . . . ) are not really confronted with a bookie who tries to make them suffer. Rather, the agents may have no idea at all what distribution holds, except that it is in some set $\mathcal{P}$. Because all they know is $\mathcal{P}$, they decide to prepare themselves for the worst-case and play the minimax strategy. The fact that such a minimax strategy can be interpreted in terms of a Nash equilibrium of a game helps to understand differences between different forms of minimax (such as a priori and a posteriori minimax). From this point of view, it seems strange to have a bookie choose between different distributions in $\mathcal{P}$ according to some distribution $\pi^*$. However, if $\mathcal{P}$ is convex, we can replace the distribution $\pi^*$ on $\mathcal{P}$ by a single distribution in $\mathcal{P}$, which consists of the convex combination of the distributions in the support of $\pi^*$; this is just the distribution $\mathrm{Pr}^*$ of Theorems 3.1 and 3.2. Thus, Theorems 3.1 and 3.2 hold with the bookie restricted to a deterministic strategy.

## 4. Conditioning, Rectangularity, and Time Consistency

To get the a posteriori minimax-optimal decision rule we do the obvious thing: if $x$ is observed, we simply condition each probability distribution $\mathrm{Pr} \in \mathcal{P}$ on $X = x$, and choose the action that gives the least expected loss (in the worst case) with respect to $\mathcal{P} \mid X = x$.

We might expect that the a priori minimax-optimal decision rule should do the same thing. That is, it should be the decision rule that says, if $x$ is observed, then we choose the action that again gives the best result (in the worst case) with respect to $\mathcal{P} \mid X = x$. But Example 2.1 shows that this cannot be true in general, since in some cases the a priori optimal decision rule is not to condition, but to ignore the observed value of $X$, and just choose the action that gives the least expected loss (in the worst case) with respect to $\mathcal{P}$, no matter what value $X$ has. We later show that there are cases in which the optimal a priori rule is neither to condition nor to ignore (see Example 5.4). Our goal in this section is to show that the rectangularity condition of Epstein and Schneider (2003) suffices to guarantee that conditioning is optimal.

**Definition 4.1:** Let $\langle \mathcal{P} \rangle$, the *hull* of $\mathcal{P}$, be the set

$$\{\Pr \in \Delta(\mathcal{X} \times \mathcal{Y}) : \Pr_{\mathcal{X}} \in \mathcal{P}_{\mathcal{X}} \text{ and, if } \Pr(X = x) \neq 0, \text{ then } (\Pr \mid X = x) \in (\mathcal{P} \mid X = x)\}.$$

∎

Thus, $\langle \mathcal{P} \rangle$ consists of all distributions $\Pr$ whose marginal on $\mathcal{X}$ is the marginal on $\mathcal{X}$ of some distribution in $\mathcal{P}$ and whose conditional on observing $X = x$ is the conditional of some distribution in $\mathcal{P}$, for all $x \in \mathcal{X}$. Clearly $\mathcal{P} \subseteq \langle \mathcal{P} \rangle$, but the converse is not necessarily true, as the following example shows.

**Example 4.2:** Suppose that $\mathcal{X} = \mathcal{Y} = \{0, 1\}$, and $\Pr_1, \Pr_2, \Pr_3 \in \Delta(\mathcal{X} \times \mathcal{Y})$ are defined as follows:

- $\Pr_1(0, 0) = \Pr_1(1, 0) = 1/3$; $\Pr_1(0, 1) = \Pr_1(1, 1) = 1/6$;

- $\Pr_2(0, 0) = \Pr_2(1, 0) = 1/6$; $\Pr_2(0, 1) = \Pr_2(1, 1) = 1/3$;

- $\Pr_3(0, 0) = \Pr_3(1, 1) = 1/3$; $\Pr_3(0, 1) = \Pr_3(1, 0) = 1/6$.

Suppose that $\mathcal{P} = \{\Pr_1, \Pr_2\}$. Then $\Pr_3 \notin \mathcal{P}$, but it is easy to see that $\Pr_3 \in \langle \mathcal{P} \rangle$. For $(\Pr_1)_{\mathcal{X}} = (\Pr_2)_{\mathcal{X}} = (\Pr_3)_{\mathcal{X}}$ is the uniform distribution on $\mathcal{X}$, $\Pr_3 \mid (X = 0) = \Pr_1 \mid (X = 0)$, and $\Pr_3 \mid (X = 1) = \Pr_2 \mid (X = 1)$. ∎

Note also that for the $\mathcal{P}$ in Example 2.1, we have $\langle \mathcal{P} \rangle = \Delta(\mathcal{X} \times \mathcal{Y}) \neq \mathcal{P}$. The notion of the hull arises in a number of contexts. In the language of Walley (1991), the hull of $\mathcal{P}$ is the *natural extension* of the marginals $\mathcal{P}_{\mathcal{X}}$ and the collection of sets of conditional probabilities $\mathcal{P} \mid X = x$ for $x \in \mathcal{X}$. Thus, if $\mathcal{P} = \langle \mathcal{P} \rangle$, then we can reconstruct the joint probability distributions $\mathcal{P}$ from $\mathcal{P}_{\mathcal{X}}$ and the collection of sets of conditional probabilities. The assumption that $\mathcal{P} = \langle \mathcal{P} \rangle$ is closely related to a set of probabilities being *separately specified*, introduced by da Rocha and Cozman (2002). As da Rocha and Cozman point out, this assumption makes it possible to apply ideas from Bayesian networks to uncertainty represented by sets of probability distributions.

The condition $\mathcal{P} = \langle \mathcal{P} \rangle$ is an instance of the *rectangularity condition* which goes back at least to the work of Sarin and Wakker (1998). It was introduced in its most general form by Epstein and Schneider (2003). Epstein and Schneider define this condition for a sequence of random variables $X_1, \ldots, X_t$, where the support of each $X_j$ is not necessarily finite. In the special case that $t = 2$, and $X := X_1$ and $Y := X_2$ are restricted to have finite support, the rectangularity condition is exactly equivalent to our condition that $\mathcal{P} = \langle \mathcal{P} \rangle$.

Considering $\langle \mathcal{P} \rangle$ also gives some insight into the two games that we considered in Section 3. In the $\mathcal{P}$-$\mathcal{X}$-game, the bookie has more power than in the $\mathcal{P}$-game, since he gets to choose the distribution after the agent observes $x$ in the $\mathcal{P}$-$\mathcal{X}$-game, and must choose it before the agent observes $x$ in the $\mathcal{P}$-game. That means that the agent can draw inferences about the distribution that the bookie chose in the $\mathcal{P}$-game. Such inferences cannot be drawn if $\mathcal{P} = \langle \mathcal{P} \rangle$. More generally, in a precise sense, the agent has the same information about $Y$ in the $\mathcal{P}$-$\mathcal{X}$-game as in the $\langle \mathcal{P} \rangle$-game. Rather than making this formal (since it is somewhat tangential to our main concerns), we give an example to show the intuition.

**Example 4.3:** Suppose that $\mathcal{X} = \mathcal{Y} = \{0, 1\}$, and $\mathcal{P} = \{\mathrm{Pr}_1, \mathrm{Pr}_2\}$, where

- $\mathrm{Pr}_1(0,0) = \epsilon(1-\epsilon)$, $\mathrm{Pr}_1(0,1) = (1-\epsilon)^2$, $\mathrm{Pr}_1(1,0) = \epsilon(1-\epsilon)$, and $\mathrm{Pr}_1(1,1) = \epsilon^2$;

- $\mathrm{Pr}_2(0,0) = \epsilon(1-\epsilon)$, and $\mathrm{Pr}_2(0,1) = \epsilon^2$, $\mathrm{Pr}_2(1,0) = \epsilon(1-\epsilon)$, $\mathrm{Pr}_2(1,1) = (1-\epsilon)^2$.

In the $\mathcal{P}$-game, if the agent observes that $X = 0$, then he is almost certain that the bookie chose $\mathrm{Pr}_1$, and thus is almost certain that $Y = 1$. On the other hand, in the $\mathcal{P}$-$X$-game, when the agent observes $x$, he has no idea whether the bookie will choose $\mathrm{Pr}_1$ or $\mathrm{Pr}_2$ (since the bookie makes this choice after observing $x$), and has no idea whether $Y$ is 0 or 1. Note that $\mathcal{P} \neq \langle \mathcal{P} \rangle$; in particular, there is a distribution $\mathrm{Pr}_3 \in \langle \mathcal{P} \rangle$ such that $(\mathrm{Pr}_3)_{\mathcal{X}} = (\mathrm{Pr}_1)_{\mathcal{X}}$ and $(\mathrm{Pr}_3) \mid (X = 0) = (\mathrm{Pr}_2) \mid (X = 0)$. For example, we can take $\mathrm{Pr}_3$ such that $\mathrm{Pr}_3(0,0) = (1-\epsilon)^2$ and $\mathrm{Pr}_3(0,1) = \epsilon(1-\epsilon)$ (the values of $\mathrm{Pr}_3(1,0)$ and $\mathrm{Pr}_3(1,1)$ are irrelevant, as long as they sum to $\epsilon$ and are nonnegative). Thus, after observing that $X = 0$ in the $\langle \mathcal{P} \rangle$ game, the agent would have no more of an idea of the value of $Y$ than he does in the $\mathcal{P}$-$\mathcal{X}$ game. ∎

The key point for us here is that when $\mathcal{P} = \langle \mathcal{P} \rangle$, conditioning is optimal, as the following theorem shows. We first need a definition. We call $\mathcal{P}$ *conservative* if for all $\mathrm{Pr} \in \mathcal{P}$ and all $x \in \mathcal{X}$, $\mathrm{Pr}(X = x) > 0$.[7]

**Theorem 4.4:** *Given a decision setting $DS = (\mathcal{X}, \mathcal{Y}, \mathcal{A}, \mathcal{P})$ such that $\mathcal{P} = \langle \mathcal{P} \rangle$, then for all decision problems $DP$ based on $DS$, there exists an a priori minimax-optimal rule that is also a posteriori minimax optimal. Indeed, every a posteriori minimax-optimal rule is also a priori minimax optimal, so $DS$ and $DP$ are weakly time consistent. Moreover, if $\mathcal{P}$ is conservative, then for every decision problem $DP$ based on $DS$, every a priori minimax-optimal rule is also a posteriori minimax optimal, so $DS$ and $DP$ are time consistent.*

This raises the question as to whether the qualification "there exists" in Theorem 4.4 is necessary, and whether the converse of the theorem also holds. Example 4.5 shows that the answer to the first question is yes; Example 4.6 shows that the answer to the second question is no.

**Example 4.5:** If for some $x \in \mathcal{X}$, there exist $\mathrm{Pr}, \mathrm{Pr}' \in \mathcal{P}$ such that $\mathrm{Pr}(X = x) = 0$ and $\mathrm{Pr}'(X = x) > 0$, then there may be an a priori minimax decision rule that is not a posteriori minimax. For example, consider the decision problem $DP = (\mathcal{X}, \mathcal{Y}, \mathcal{A}, \mathcal{P}, L)$ with

---

7. Our notion of conservative corresponds to what Epstein and Schneider (2003) call the *full support* condition.

$\mathcal{X} = \{0, 1\}$, $\mathcal{A} = \mathcal{Y} = \{0, 1, 2\}$, $L$ the classification loss (Example 2.1) and $\mathcal{P} = \{\text{Pr}_1, \text{Pr}_2\}$. We first define $\text{Pr}_1$:

$\text{Pr}_1(X = 1) = 1/2,$
$\text{Pr}_1(Y = 0 \mid X = 0) = \text{Pr}_1(Y = 1 \mid X = 0) = \text{Pr}_1(Y = 2 \mid X = 0) = 1/3,$ and
$\text{Pr}_1(Y = 0 \mid X = 1) = 1/2,$
$\text{Pr}_1(Y = 1 \mid X = 1) = 2/5,$
$\text{Pr}_1(Y = 2 \mid X = 1) = 1/10.$

$\text{Pr}_2$ is defined as follows: $\text{Pr}_2(X = 0) = 1$, and for all $j \in \mathcal{Y}$, $\text{Pr}_2(Y = j, X = 0) = \text{Pr}_2(Y = j \mid X = 0) := \text{Pr}_1(Y = j \mid X = 0)$. It is easy to see that $\mathcal{P} = \langle \mathcal{P} \rangle$, so the rectangularity condition holds.

Note that $\delta(0)$, the decision taken when observing $X = 0$, does not affect the expected loss; for both $\text{Pr}_1 \mid X = 0$ and $\text{Pr}_2 \mid X = 0$, $Y$ is uniform, so the expected loss is $2/3$, regardless of $\delta(0)$. This implies that every decision rule $\delta$ with $\delta(1)$ a randomized combination of $\{0, 1\}$ is a priori optimal, and has worst-case expected loss $2/3$, since $E_{\text{Pr}_2}[L_\delta] = 2/3$ and $E_{\text{Pr}_1}[L_\delta] < 2/3$. But the minimax optimal rules with $\delta(1) = 1$ are not a posteriori optimal, since if the player observes $X = 1$, he knows that the distribution is $\text{Pr}_1$, and the minimax loss relative to $\text{Pr}_1$ is $1/2$ for action 0 and $3/5$ for action 1.

Both in this example and in Example 4.3, observing a particular value of $X$ gives information about which distribution in $\mathcal{P}$ the bookie has chosen. In Example 4.3, observing $X = 0$ implies that the bookie almost certainly chose $\text{Pr}_1$ in the $\mathcal{P}$-game; in the present example, observing $X = 1$ implies that the bookie certainly chose $\text{Pr}_1$ in both the $\mathcal{P}$-game and the $\mathcal{P} - X$ game. We note, however, that observing $X = x$ can give information about the distribution chosen by the bookie in the $\mathcal{P} - X$ game *only* if there exist $\text{Pr}$ and $\text{Pr}'$ in the $\mathcal{P}$-game such that $\text{Pr}(X = x) = 0$ and $\text{Pr}'(X = x) > 0$. If no such $\text{Pr}$ and $\text{Pr}'$ exists, then the bookie is completely free to choose any $\text{Pr} \in \mathcal{P}$ he likes after $x$ has been observed, so observing $x$ gives no information about which $\text{Pr} \in \mathcal{P}$ has been chosen. ■

There exist decision settings such that $\mathcal{P}$ is conservative and $\mathcal{P} \neq \langle \mathcal{P} \rangle$, although we still have weak time consistency. Hence, the converse of Theorem 4.4 does not hold in general. We now give an example of such a $\mathcal{P}$.

**Example 4.6:** Let $\mathcal{X} = \mathcal{A} = \mathcal{Y} = \{0, 1\}$ and $\mathcal{P} = \{\text{Pr}_0, \text{Pr}_1\}$ with $\text{Pr}_0(X = 1) = \text{Pr}_1(X = 1) = 1/2$ and for $x \in \{0, 1\}$, $\text{Pr}_0(Y = 0 \mid X = x) = 1$ and $\text{Pr}_1(Y = 1 \mid X = x) = 1$. Clearly $\mathcal{P}$ is conservative and $\mathcal{P} \neq \langle \mathcal{P} \rangle$; for example, the distribution $\text{Pr}_3$ such that $\text{Pr}_3(X = 1) = 1/2$, $\text{Pr}_3(Y = 0 \mid X = 0) = 1$, and $\text{Pr}_3(Y = 0 \mid X = 1) = 0$ is in $\langle \mathcal{P} \rangle - \mathcal{P}$. Note that $X$ and $Y$ are independent with respect to both $\text{Pr}_0$ and $\text{Pr}_1$. Now take an arbitrary loss function $L$. Since $(\text{Pr} \mid X = x)_{\mathcal{Y}}$ contains two distributions, one with $\text{Pr}(Y = 1) = 0$ and one with $\text{Pr}(Y = 1) = 1$, the minimax a posteriori act is to play $\delta(0) = \delta(1) = (1 - \alpha^*) \cdot 0 + \alpha^* \cdot 1$ (i.e., the act that plays 0 with probability $1 - \alpha^*$ and 1 with probability $\alpha^*$), where $\alpha^*$ is chosen so as to minimize $f(\alpha) = \max\{(1 - \alpha)L(0, 0) + \alpha L(0, 1), (1 - \alpha)L(1, 0) + \alpha L(1, 1)\}$. For simplicity, assume that there is a unique such $\alpha^*$. (If not, then it must be the case that all $\alpha \in [0, 1]$ minimize this expression, and it is easy to check $L(0, 0) = L(0, 1) = L(1, 0) = L(1, 1)$, so time consistency holds trivially.)

405

We want to show that $\delta$ is also a priori minimax. It is easy to check that

$$\max_{\Pr \in \{\Pr_0, \Pr_1\}} L_\delta = f(\alpha^*),$$

where $f$ is as above. So it suffices to show that for any decision rule $\delta'$, we must have

$$\max_{\Pr \in \{\Pr_0, \Pr_1\}} L_{\delta'} \geq f(\alpha^*),$$

Suppose that $\delta(x) = (1 - \beta_x) \cdot 0 + \beta_x \cdot 1$, for $x \in \{0, 1\}$. Then

$$
\begin{aligned}
&\max_{\Pr \in \{\Pr_0, \Pr_1\}} E_{Pr}[L_{\delta'}] \\
= {}&\max\{\tfrac{1}{2}((1 - \beta_0)L(0,0) + \beta_0 L(0,1) + (1 - \beta_1)L(0,0) + \beta_1 L(0,1)), \\
&\qquad \tfrac{1}{2}((1 - \beta_0)L(1,0) + \beta_0 L(1,1) + (1 - \beta_1)L(1,0) + \beta_1 L(1,1)\} \\
= {}&\max\{(1 - \gamma)L(0,0) + \gamma L(0,1), (1 - \gamma)L(1,0) + \gamma L(1,1)\}, \text{ where } \gamma = \tfrac{\beta_0 + \beta_1}{2} \\
= {}&f(\gamma) \geq f(\alpha^*).
\end{aligned}
$$

∎

It is interesting to compare Theorem 4.4 with the results of Epstein and Schneider (2003). For this, we first compare our notion of time consistency with their notion of dynamic consistency. Both notions were formally defined at the end of Section 2. Our results are summarized in Proposition 4.7. First we need two definitions: Let $\mathcal{P}$ be a set of distributions on $\mathcal{X} \times \mathcal{Y}$. A decision problem is *based on* $\mathcal{P}$ if it is of the form $(\mathcal{X}, \mathcal{Y}, \mathcal{A}, \mathcal{P}, L)$ for some arbitrary $\mathcal{A}$ and $L$. A decision problem satisfies *strong dynamic consistency* if it satisfies condition (2) of the definition of dynamic consistency and satisfies the following strengthening of (3):

- If, for all $x$ such that $\Pr(X = x) > 0$ for some $\Pr \in \mathcal{P}$, (2) holds, and for some $x$ such that $\Pr(X = x) > 0$, we have

$$\max_{\Pr \in (\mathcal{P} | X = x)} E_{\Pr}[L_\delta] < \max_{\Pr \in (\mathcal{P} | X = x)} E_{\Pr}[L_{\delta'}], \tag{4}$$

then (3) must hold with strict inequality.

**Proposition 4.7:**

(a) *Every dynamically consistent decision problem is also weakly time consistent.*

(b) *Not every dynamically consistent decision problem is time consistent.*

(c) *Every strongly dynamically consistent decision problem is time consistent.*

(d) *There exist weakly time consistent decision problems that are not dynamically consistent.*

(e) *All decision problems based on $\mathcal{P}$ are dynamically consistent if and only if all decision problems based on $\mathcal{P}$ are weakly time consistent.*

Proposition 4.7(c) shows that the comparison between time consistency and dynamic consistency is subtle: replacing 'for all $x$' by "for some $x$' in the second half of the definition of dynamic consistency, which leads to a perfectly reasonable requirement, suffices to force time consistency. Proposition 4.7(e) leads us to suspect that a decision setting is weakly time consistent if and only if it is dynamically consistent. We have, however, no proof of this claim. The proof of part (e) involves two decision problems based on the same set $\mathcal{P}$, but with different sets of actions, so these decision problems are not based on the same decision setting. It does not seem straightforward to extend the result to decision settings.

Epstein and Schneider show, among other things, that if $\mathcal{P}$ is closed, convex, conservative, and rectangular, then $DS$ is is dynamically consistent, and hence weakly time consistent. We remark that the convexity assumption is not needed for this result. It easy to check that $\delta$ is prefered to $\delta'$ with respect to $\mathcal{P}$ according to the minimax criterion iff $\delta$ is preferred to $\delta'$ with respect to the convex closure of $\mathcal{P}$ according to the minimax criterion. Proposition 4.7 shows that dynamic and time consistency are closely related. Yet, while there is clear overlap in what we prove in Theorem 4.4 and the Epstein-Schneider (ES from now on) result, in general the results are incomparable. For example, we can already prove weak time consistency without assuming conservativeness; ES assume conservativeness throughout. On the other hand, ES also show that if dynamic consistency holds, then the agent's actions can be viewed as being the minimax optimal actions relative to a rectangular convex conservative set; we have no analogous result for time consistency. Moreover, in contrast to the ES result, our results hold only for the restricted setting with just two time steps, one before and one after making a single observation.

## 5. Belief Updates and $\mathcal{C}$-conditioning

In this section we define the notion of a belief update rule, when belief is represented by sets of probabilities, and introduce a natural family of belief update rules which we call $\mathcal{C}$-conditioning.

To motivate these notions, recall that Example 2.1 shows that the minimax-optimal a priori decision rule is not always the same as the minimax-optimal a posteriori decision rule. In this example, the minimax-optimal a priori decision rule ignores the information observed. Formally, a rule $\delta$ *ignores information* if $\delta(x) = \delta(x')$ for all $x, x' \in \mathcal{X}$. If $\delta$ ignores information, define $L'_\delta$ to be the random variable on $\mathcal{Y}$ such that $L'_\delta(y) = L_\delta(x, y)$ for some choice of $x$. This is well defined, since $L_\delta(x, y) = L_\delta(x', y)$ for all $x, x' \in \mathcal{X}$.

The following theorem provides a general sufficient condition for ignoring information to be optimal.

**Theorem 5.1:** *Fix $\mathcal{X}$, $\mathcal{Y}$, $L$, $\mathcal{A}$, and $\mathcal{P} \subseteq \Delta(\mathcal{X} \times \mathcal{Y})$. If, for all $\Pr_\mathcal{Y} \in \mathcal{P}_\mathcal{Y}$, $\mathcal{P}$ contains a distribution $\Pr'$ such that $X$ and $Y$ are independent under $\Pr'$, and $\Pr'_\mathcal{Y} = \Pr_\mathcal{Y}$, then there is an a priori minimax-optimal decision rule that ignores information. Under these conditions, if $\delta$ is an a priori minimax-optimal decision rule that ignores information, then $\delta$ essentially optimizes with respect to the marginal on $Y$; that is, $\max_{\Pr \in \mathcal{P}} E_{\Pr}[L_\delta] = \max_{\Pr_\mathcal{Y} \in \mathcal{P}_\mathcal{Y}} E_{\Pr_\mathcal{Y}}[L'_\delta]$.*

GH focused on the case that $\mathcal{P}_{\mathcal{Y}}$ is a singleton (i.e., the marginal probability on $Y$ is the same for all distributions in $\mathcal{P}$) and for all $x$, $\mathcal{P}_{\mathcal{Y}} \subseteq (\mathcal{P} \mid X = x)_{\mathcal{Y}}$. It is immediate from Theorem 5.1 that ignoring information is a priori minimax optimal in this case.

Standard conditioning and ignoring information are both instances of $\mathcal{C}$-conditioning, which in turn is an instance of an update rule. We now define these notions formally.

**Definition 5.2:** A *belief update rule* (or just an *update rule*) is a function $\Pi : 2^{\Delta(\mathcal{X} \times \mathcal{Y})} \times \mathcal{X} \rightarrow 2^{\Delta(\mathcal{X} \times \mathcal{Y})} - \{\emptyset\}$ mapping a set $\mathcal{P}$ of distributions and an observation $x$ to a nonempty set $\Pi(\mathcal{P}, x)$ of distributions; intuitively, $\Pi(\mathcal{P}, x)$ is the result of updating $\mathcal{P}$ with the observation $x$. ∎

In the case where $\mathcal{P}$ is a singleton $\{\text{Pr}\}$, then one update rule is conditioning; that is, $\Pi(\{\text{Pr}\}, x) = \{\text{Pr}(\cdot \mid X = x)\}$. But other update rules are possible, even for a single distribution; for example, Lewis (1976) considered an approach to updating that he called *imaging*. There is even more scope when considering sets of probabilities; for example, both Walley's (1991) natural extension and regular extension provide update rules (as we said, our notion of conditioning can be viewed as an instance of Walley's regular extension). Simply ignoring information provides another update rule: $\Pi(\mathcal{P}, x) = \mathcal{P}$. As we said above, ignoring information and standard conditioning are both instances of $\mathcal{C}$-conditioning.

**Definition 5.3:** Let $\mathcal{C} = \{\mathcal{X}_1, \ldots, \mathcal{X}_k\}$ be a partition of $\mathcal{X}$; that is, $\mathcal{X}_i \neq \emptyset$ for $i = 1, \ldots, k$; $\mathcal{X}_1 \cup \ldots \cup \mathcal{X}_k = \mathcal{X}$; and $\mathcal{X}_i \cap \mathcal{X}_j = \emptyset$ for $i \neq j$. If $x \in \mathcal{X}$, let $\mathcal{C}(x)$ be the cell containing $x$; that is, the unique element $\mathcal{X}_i \in \mathcal{C}$ such that $x \in \mathcal{X}_i$. The $\mathcal{C}$-*conditioning* belief update rule is the function $\Pi$ defined by taking $\Pi(\mathcal{P}, x) = \mathcal{P} \mid \mathcal{C}(x)$ (if for all $\text{Pr} \in \mathcal{P}$, $\text{Pr}(\mathcal{C}(x)) = 0$, then $\Pi(\mathcal{P}, x)$ is undefined). A decision rule $\delta$ is *based on $\mathcal{C}$-conditioning* if it amounts to first updating the set $\mathcal{P}$ to $\mathcal{P} \mid \mathcal{C}(x)$, and then taking the minimax-optimal distribution over actions relative to $(\mathcal{P} \mid \mathcal{C}(x))_{\mathcal{Y}}$. Formally, $\delta$ is based on $\mathcal{C}$-conditioning if, for all $x \in \mathcal{X}$ with $\text{Pr}(X = x) > 0$ for some $\text{Pr} \in \mathcal{P}$,

$$\max_{\text{Pr} \in (\mathcal{P} \mid X \in \mathcal{C}(x))_{\mathcal{Y}}} E_{\text{Pr}}[L_{\delta(x)}] = \min_{\gamma \in \Delta(\mathcal{A})} \max_{\text{Pr} \in (\mathcal{P} \mid X \in \mathcal{C}(x))_{\mathcal{Y}}} E_{\text{Pr}}[L_\gamma].$$

∎

Standard conditioning is a special case of $\mathcal{C}$-conditioning, where we take $\mathcal{C}$ to consist of all singletons; ignoring information is also based on $\mathcal{C}$-conditioning, where $\mathcal{C} = \{\mathcal{X}\}$. Our earlier results suggest that perhaps an a priori minimax-optimal decision rule must be based on $\mathcal{C}$-conditioning for some $\mathcal{C}$. The Monty Hall problem again shows that this conjecture is false.

**Example 5.4: [Monty Hall]** (Mosteller, 1965; vos Savant, 1990): We start with the original Monty Hall problem, and then consider a variant of it. Suppose that you're on a game show and given a choice of three doors. Behind one is a car; behind the others are goats. You pick door 1. Before opening door 1, Monty Hall, the host (who knows what is behind each door) opens one of the other two doors, say, door 3, which has a goat. He then asks you if you still want to take what's behind door 1, or to take what's behind door 2 instead. Should you switch? You may assume that initially, the car was equally likely to be behind each of the doors.

We formalize this well-known problem as a $\mathcal{P}$-game, as follows: $\mathcal{Y} = \{1, 2, 3\}$ represents the door which the car is behind. $\mathcal{X} = \{G_2, G_3\}$, where, for $j \in \{2, 3\}$, $G_j$ corresponds to the quizmaster showing that there is a goat behind door $j$. $\mathcal{A} = \{1, 2, 3\}$, where action $a \in \mathcal{A}$ corresponds to the door you finally choose, after Monty has opened door 2 or 3. The loss function is once again the classification loss, $L(i, j) = 1$ if $i \neq j$, that is, if you choose a door with a goat behind it, and $L(i, j) = 0$ if $i = j$, that is, if you choose a door with a car. $\mathcal{P}$ is the set of all distributions Pr on $\mathcal{X} \times \mathcal{Y}$ satisfying

$$\mathrm{Pr}_{\mathcal{Y}}(Y = 1) = \mathrm{Pr}_{\mathcal{Y}}(Y = 2) = \mathrm{Pr}_{\mathcal{Y}}(Y = 3) = \tfrac{1}{3}$$
$$\mathrm{Pr}(Y = 2 \mid X = G_2) = \mathrm{Pr}(Y = 3 \mid X = G_3) = 0.$$

Note that $\mathcal{P}$ does not satisfy the rectangularity condition. For example, let $\mathrm{Pr}^*$ be the distribution such that $\mathrm{Pr}^*(G_2, 1) = \mathrm{Pr}^*(G_2, 3) = 1/3$ and $\mathrm{Pr}^*(G_3, 1) = \mathrm{Pr}^*(G_3, 2) = 1/6$. It is easy to see that $\mathrm{Pr}^* \in \langle \mathcal{P} \rangle - \mathcal{P}$.

It is well known, and easy to show, that the a priori minimax-optimal strategy is always to switch doors, no matter whether Monty opens door 2 or door 3. Formally, let $\delta_S$ be the decision rule such that $\delta_S(G_2) = 3$ and $\delta_S(G_3) = 2$. Then $\delta_S$ is the unique a priori minimax-optimal decision rule (and has expected loss 1/3). The rule $\delta_S$ is also a posteriori minimax optimal. But now we modify the problem so that there is a small cost, say $\epsilon > 0$, associated with switching. The cost is associated both with switching to door 2 and with switching to door 3. As long as $\epsilon$ is sufficiently small, the action $\delta_S$ of always switching is still uniquely a priori minimax optimal. However, now $\delta_S$ is *not* based on $\mathcal{C}$-conditioning. There exist only two partitions of $\mathcal{X}$. The corresponding two update rules based on $\mathcal{C}$-conditioning amount to, respectively, (1) ignoring $X$, and (2) conditioning on $X$ in the standard way. The decision rule based on ignoring the information is to stick to door 1, because there is a cost associated with switching. The decision rule based on conditioning is to switch doors with probability $1/(2 + \epsilon)$. To see this, consider the observation $X = G_2$, and let $\alpha$ be the randomized action of switching to door 3 with probability $q$ and sticking to door 1 with probability $1 - q$. Let $m(q) = \max_{\mathrm{Pr} \in \langle \mathcal{P} \mid X = G_2 \rangle_{\mathcal{Y}}} E_{\mathrm{Pr}}[L_\alpha]$. Thus, $m(q) = \max_{p \in [0, 1/2]} (qp(1 + \epsilon) + (1 - q)(1 - p))$. Again, to compute $m(q)$, we need to consider only what happens when is at the extremes of the interval; that is, when $p = 0$ or $p = 1/2$, so $m(q) = \max(1 - q, (1 + q\epsilon)/2)$. Clearly $m(q)$ is minimized when $1 - q = (1 + q\epsilon)/2$, that is, when $q = 1/(2 + \epsilon)$. A similar analysis applies when the observation $X = G_3$. Thus, neither of the decision rules based on conditioning is minimax optimal. ∎

Although $\mathcal{C}$-conditioning does not guarantee minimax optimality, it turns out to be a useful notion. As we show in the next section, it is quite relevant when we consider calibration.

## 6. Calibration

As we said in the introduction, Dawid (1982) pointed out that an agent who is updating his beliefs should want to be calibrated. In this section, we consider the effect of requiring calibration. Up to now, calibration has been considered only when uncertainty is characterized by a single distribution. Below we generalize the notion of calibration to our setting, where uncertainty is characterized by a set of distributions. We then investigate the connection between calibration and some of the other conditions that we considered earlier, specifically the conditions that $\mathcal{P}$ is convex and $\mathcal{P} = \langle \mathcal{P} \rangle$.[8]

Calibration is typically defined with respect to empirical data. We view the set $\mathcal{P}$ of distributions not as describing empirical data, but as defining an agent's uncertainty regarding the true distribution. We want to define calibration in such a setting. For the case that $\mathcal{P}$ is a singleton, this has already been done, for example, by Vovk, Gammerman, and Shafer (2005). [9] Below, we first define calibration for the case where $\mathcal{P}$ is a singleton, and then extend the notion to general $\mathcal{P}$.

Let $\Pi$ be an update rule such that $\Pi(\{\Pr\}, x)$ contains just a single distribution for each $x \in \mathcal{X}$ (for example, $\Pi$ could be ordinary conditioning). Given $x \in \mathcal{X}$ and $\Pi$, define $[x]_{\Pi, \mathcal{P}} = \{x' : (\Pi(\mathcal{P}, x'))_{\mathcal{Y}} = \Pi(\mathcal{P}, x)_{\mathcal{Y}}\}$. Thus, $[x]_{\Pi, \mathcal{P}}$ consists of all values $x'$ that, when observed, lead to the same updated marginal distributions as $x$.

**Definition 6.1:** The update rule $\Pi$ is *calibrated relative to* $\Pr$ if, for all $x \in \mathcal{X}$, if $\Pr([x]_{\Pi, \{\Pr\}}) \neq 0$, then $\Pr(\cdot \mid [x]_{\Pi, \{\Pr\}})_{\mathcal{Y}} = \Pi(\{\Pr\}, x)_{\mathcal{Y}}$.[10] ∎

In words, this definition says that if $\Pr'$ is the distribution on $\mathcal{Y}$ that results from updating $\Pr$ after observing $x$ according to $\Pi$ and then marginalizing to $\mathcal{Y}$, then $\Pi$ is calibrated if $\Pr'$ is also the marginal distribution that results when conditioning $\Pr$ on the set of values $x'$ that, when observed, result in $\Pr'$ being the marginal distribution according to $\Pi$. Intuitively, for each $x$ that may be observed, an agent who uses $\Pi$ produces a distribution $\Pi(\{\Pr\}, x)$. The agent may then make decisions or predictions about $Y$ based on this distribution, marginalized to $\mathcal{Y}$. We consider the set $\mathcal{P}'$ of all distributions on $\mathcal{Y}$ that the agent may use to predict $Y$ after observing the value of $X$. That is, $\Pr' \in \mathcal{P}'$ iff with positive $\Pr$-probability the agent, after observing the value of $X$, uses $\Pr'$ to predict $Y$. The set $\mathcal{P}'$ has at most $|\mathcal{X}|$ elements. Definition 6.1 then says that, for each $\Pr' \in \mathcal{P}'$, whenever the agent predicts with $\Pr'$, the agent is "correct" in the sense that the distribution of $\mathcal{Y}$ given that the agent uses $\Pr'$ is indeed to $\Pr'$. Note that in Definition 6.1, as in all subsequent definitions in this section, we marginalize on $\mathcal{Y}$. We discuss this further at the end of this section. It is straightforward to generalize Definition 6.1 to sets $\mathcal{P}$ of probability distributions that are not singletons, and update rules $\Pi$ that map to sets of probabilities.

**Definition 6.2:** The update rule $\Pi$ is *calibrated relative to* $\mathcal{P}$ if, for all $x \in \mathcal{X}$, if $\Pr([x]_{\Pi, \mathcal{P}}) \neq 0$ for some $\Pr \in \mathcal{P}$, then $(\mathcal{P} \mid [x]_{\Pi, \mathcal{P}})_{\mathcal{Y}} = \Pi(\mathcal{P}, x)_{\mathcal{Y}}$. ∎

We now want to relate calibration and $\mathcal{C}$-conditioning. The following result is a first step in that direction. It gives conditions under which standard conditioning is calibrated, and also shows that, for convex $\mathcal{P}$ and arbitrary $\mathcal{C}$, $\mathcal{C}$-conditioning satisfies one of the two inclusions required by Definition 6.2.

---

8. Recall that convexity is an innocuous assumption in the context of time and dynamic consistency. However, as we show in this section, it is far from innocuous in the context of calibration.

9. Vovk et al.'s setting is somewhat different from ours, because they are interested only in upper bounds on, rather than precise values of, probabilities. As a result, their definition of "validity" (as they call their notion of calibration) is somewhat different from Definition 6.1, but the underlying idea is the same. We have found no definition in the literature that coincides with ours.

10. As usual, if $A \subseteq \mathcal{X}$, then we identify $\mathcal{P} \mid A$ with $\mathcal{P} \mid (A \times \mathcal{Y})$.

**Theorem 6.3:**

    (a) If $\Pi$ is $\mathcal{C}$-conditioning for some partition $\mathcal{C}$ of $\mathcal{X}$ and $\mathcal{P}$ is convex then, for all $x \in \mathcal{X}$, we have that $(\mathcal{P} \mid [x]_{\Pi,\mathcal{P}})_{\mathcal{Y}} \subseteq \Pi(\mathcal{P}, x)_{\mathcal{Y}}$.

    (b) If $\Pi$ is standard conditioning, $\mathcal{P} = \langle \mathcal{P} \rangle$, and $x \in \mathcal{X}$, then $\Pi(\mathcal{P}, x)_{\mathcal{Y}} \subseteq (\mathcal{P} \mid [x]_{\Pi,\mathcal{P}})_{\mathcal{Y}}$.

**Corollary 6.4:** *If $\mathcal{P}$ is convex and $\mathcal{P} = \langle \mathcal{P} \rangle$, then standard conditioning is calibrated relative to $\mathcal{P}$.*

This corollary will be significantly strengthened in Theorem 6.12 below. In general, both convexity and the $\mathcal{P} = \langle \mathcal{P} \rangle$ condition are necessary in Corollary 6.4, as the following two examples show.

**Example 6.5:** Let $\mathcal{X} = \mathcal{Y} = \{0, 1\}$, let $\mathcal{P} = \{\Pr_1, \Pr_2, \Pr_3, \Pr_4\}$, where $\Pr_1, \ldots, \Pr_4$ are defined below as a sequence of four numbers $(a, b, c, d)$, with $\Pr_i(0,0) = a$, $\Pr_i(0,1) = b$, $\Pr_i(1,0) = c$, and $\Pr_i(1,1) = d$):

- $\Pr_1 = (1/4, 1/4, 1/4, 1/4)$,

- $\Pr_2 = (1/8, 3/8, 1/8, 3/8)$,

- $\Pr_3 = (1/4, 1/4, 1/8, 3/8)$,

- $\Pr_4 = (1/8, 3/8, 1/4, 1/4)$.

Clearly $\mathcal{P}$ is not convex. Note that $\Pr_1(Y = 0 \mid X = 0) = \Pr_1(Y = 0 \mid X = 1) = 1/2, \Pr_2(Y = 0 \mid X = 0) = \Pr_2(Y = 0 \mid X = 1) = 1/4$, and $\Pr_3(Y = 0 \mid X = 0) = 1/2$, $\Pr_3(Y = 0 \mid X = 1) = 1/4$. Since, for all $\Pr \in \mathcal{P}$, $\Pr(X = 0) = 1/2$, and $(\mathcal{P} \mid X = 0)_{\mathcal{Y}} = (\mathcal{P} \mid X = 1)_{\mathcal{Y}} = \{\Pr_a, \Pr_b\}$ where $\Pr_a(Y = 0) = 1/2$ and $\Pr_b(Y = 0) = 1/4$, we have $\mathcal{P} = \langle \mathcal{P} \rangle$. We now show that standard conditioning is not calibrated relative to $\mathcal{P}$. Let $\Pi$ stand for standard conditioning. For $x \in \{0, 1\}$, we have

$$\Pi(\mathcal{P}, x)_{\mathcal{Y}} = (\mathcal{P} \mid X = x)_{\mathcal{Y}} = \{\Pr'_1, \Pr'_2\}, \tag{5}$$

where $\Pr'_1(Y = 0) = 1/2$ and $\Pr'_2(Y = 0) = 1/4$. It also follows that, for $x \in \{0, 1\}$, $[x]_{\Pi,\mathcal{P}} = \{0, 1\} = \mathcal{X}$, so that

$$(\mathcal{P} \mid [x]_{\Pi,\mathcal{P}})_{\mathcal{Y}} = \mathcal{P}_{\mathcal{Y}}. \tag{6}$$

Since $\mathcal{P}_{\mathcal{Y}}$ contains a distribution $\Pr'_3$ such that $\Pr'_3(Y = 0) = 3/8$, (5) and (6) together show that $\Pi$ is not calibrated. ∎

**Example 6.6:** Let $\mathcal{X} = \mathcal{Y} = \{0, 1\}$, and let $\mathcal{P}$ consist of all distributions on $\mathcal{X} \times \mathcal{Y}$ with $\Pr(Y = 1) = 0.5$. Clearly $\mathcal{P}$ is convex. However, $\mathcal{P} \neq \langle \mathcal{P} \rangle$. To see this, note that $\mathcal{P}$ contains a distribution $\Pr$ with $\Pr(Y = 0 \mid X = 0) = 1$ and a distribution $\Pr'$ with $\Pr'(X = 0) = 1$, but no distribution $\Pr''$ with $\Pr''(X = 0) = 1$ and $\Pr''(Y = 0 \mid X = 0) = 1$. Let $\Pi$ stand for standard conditioning. We now show that $\Pi$ is not calibrated. For $x \in \{0, 1\}$, we have

$$\Pi(\mathcal{P}, x)_{\mathcal{Y}} = (\mathcal{P} \mid X = x)_{\mathcal{Y}} = \Delta(\mathcal{Y}), \tag{7}$$

that is, conditioning both on $X = 0$ and on $X = 1$ leads to the set of all distributions on $\mathcal{Y}$. It follows that, for $x \in \{0, 1\}$, $[x]_{\Pi, \mathcal{P}} = \{0, 1\} = \mathcal{X}$, so that

$$(\mathcal{P} \mid [x]_{\Pi, \mathcal{P}})_{\mathcal{Y}} = \mathcal{P}_{\mathcal{Y}} = \{\Pr \in \Delta(\mathcal{Y}) \mid \Pr(Y = 1) = 0.5\}. \tag{8}$$

Together, (7) and (8) show that $\Pi$ is not calibrated. ∎

Corollary 6.4 gives conditions under which standard conditioning is calibrated. Theorem 6.3(a) gives general conditions under which $\mathcal{C}$-conditioning satisfies one inclusion required for calibration; specifically, $(\mathcal{P} \mid [x]_{\Pi, \mathcal{P}})_{\mathcal{Y}} \subseteq \Pi(\mathcal{P}, x)_{\mathcal{Y}}$. Rather than trying to find conditions under which the other inclusion holds, we consider a strengthening of calibration, which is arguably a more interesting notion. For, as the following example shows, calibration it is arguably too weak a requirement.

**Example 6.7:** Let $\mathcal{X} = \mathcal{Y} = \{0, 1\}$, and let $\mathcal{P} = \{\Pr\}$ consist of all distributions on $\mathcal{X} \times \mathcal{Y}$ satisfying $\Pr(Y = X) = 1$. Then the rule $\Pi$ that ignores $X$, that is, with $\Pi(\mathcal{P}, x) = \mathcal{P}$ for $x \in \{0, 1\}$, is calibrated, even though (a) it outputs all distributions on $\mathcal{Y}$, and (b) there exists another calibrated rule (standard conditioning) that, upon observing $X = x$, outputs only one distribution on $\mathcal{Y}$. ∎

Intuitively, the fewer distributions that there are in $\mathcal{P}$, the more information $\mathcal{P}$ contains. Thus, we want to restrict ourselves to sets $\mathcal{P}$ that are as small as possible, while still being calibrated.

**Definition 6.8:** Update rule $\Pi'$ is *narrower than update rule* $\Pi$ *relative to* $\mathcal{P}$ if, for all $x \in \mathcal{X}$, $\Pi'(\mathcal{P}, x)_{\mathcal{Y}} \subseteq \Pi(\mathcal{P}, x)_{\mathcal{Y}}$. $\Pi'$ is *strictly narrower* relative to $\mathcal{P}$ if the inclusion is strict for some $x$. $\Pi$ is *sharply calibrated* if there exists no update rule $\Pi'$ that is strictly narrower than $\Pi$ and that is also calibrated. ∎

We now show that if $\mathcal{P}$ is convex, then every sharply calibrated update rule must involve $\mathcal{C}$-conditioning. To make this precise, we need the following definition.

**Definition 6.9:** $\Pi$ is a *generalized conditioning update rule* if, for all convex $\mathcal{P}$, there exists a partition $\mathcal{C}$ (that may depend on $\mathcal{P}$) such that for all $x \in \mathcal{X}$, $\Pi(\mathcal{P}, x) = \mathcal{P} \mid C(x)$. ∎

Note that, as long as $\mathcal{P}$ is convex, in a generalized conditioning rule, we condition on a partition of $\mathcal{X}$, but the partition may depend on the set $\mathcal{P}$. For example, for some convex $\mathcal{P}$, the rule may ignore the value of $x$, whereas for other convex $\mathcal{P}$, it may amount to ordinary conditioning. Since we are only interested in generalized conditioning rules when $\mathcal{P}$ is convex, their behavior on nonconvex $\mathcal{P}$ is irrelevant. Indeed, the next result shows that, if we require only that $\mathcal{P}$ be convex (and do not require that $\mathcal{P} = \langle \mathcal{P} \rangle$), then $\mathcal{C}$-conditioning is calibrated, indeed, sharply calibrated, for *some* $\mathcal{C}$; moreover, every *sharply* calibrated update rule must be a generalized conditioning rule.

**Theorem 6.10:** *Suppose that* $\mathcal{P}$ *is convex.*

*(a) $\mathcal{C}$-conditioning is sharply calibrated relative to* $\mathcal{P}$ *for some partition* $\mathcal{C}$.

    *(b) If $\Pi$ is sharply calibrated relative to $\mathcal{P}$, then there exists some $\mathcal{C}$ such that $\Pi$ is equivalent to $\mathcal{C}$-conditioning on $\mathcal{P}$ (i.e., $\Pi(\mathcal{P}, x) = \Pi \mid \mathcal{C}(x)$ for all $x \in \mathcal{X}$).*

**Corollary 6.11:** *There exists a generalized conditioning update rule that is sharply calibrated relative to all convex $\mathcal{P}$. Moreover, every update rule that is sharply calibrated relative to all convex $\mathcal{P}$ is a generalized conditioning update rule relative to the set of all convex $\mathcal{P}$.*

Theorem 6.10 establishes a connection between sharp calibration and $\mathcal{C}$-conditioning. We now show that the same conditions that make standard conditioning calibrated also make it sharply calibrated.

**Theorem 6.12:** *If $\mathcal{P}$ is convex and $\mathcal{P} = \langle \mathcal{P} \rangle$, then standard conditioning is sharply calibrated relative to $\mathcal{P}$.*

This result shows that the $\mathcal{P} = \langle \mathcal{P} \rangle$ condition in Theorem 6.12 is not just relevant for ensuring time consistency, but also for ensuring the well-behavedness of conditioning in terms of calibration. Note, however, that the result says nothing about $\mathcal{C}$-conditioning for arbitrary partitions $\mathcal{C}$. In general, $\mathcal{C}$-conditioning may be sharply calibrated relative to some convex $\mathcal{P}$ with $\mathcal{P} = \langle \mathcal{P} \rangle$, but not relative to others. For example, if $\mathcal{P}$ is a singleton, then it is convex, $\mathcal{P} = \langle \mathcal{P} \rangle$, and the update rule that ignores $x$ is sharply calibrated. In Example 6.7, $\mathcal{P}$ is also convex and $\mathcal{P} = \langle \mathcal{P} \rangle$, yet ignoring $x$ is not sharply calibrated.

**Remark** All the results in this section were based on a definition of calibration in which the updated set of distributions $\Pi(\mathcal{P}, x)$ is marginalized to $\mathcal{Y}$. It is also possible to define calibration without this marginalization. However, we found that this makes for a less interesting notion. For example, without marginalizing on $\mathcal{Y}$ there no longer seems to be a straightforward way of defining "sharp" calibration, and without sharpness, the notion is of quite limited interest. Moreover, it does not seem possible to state and prove an analogue of Theorem 6.3 (at least, we do not know how to do it).

## 7. Discussion and Related Work

We have examined how to update uncertainty represented by a set of probability distributions, where we motivate updating rules in terms of the minimax criterion. Our key innovation has been to show how different approaches can be understood in terms of a game between a bookie and an agent, where the bookie picks a distribution from the set and the agent chooses an action after making an observation. Different approaches to updating arise depending on whether the bookie's choice is made before or after the observation. We believe that this game-theoretic approach should prove useful more generally in understanding different approaches to updating. In fact, after the publication of the conference version of this paper, we learned that Ozdenoren and Peck (2008) use the same type of approach for analyzing dynamic situations related to the Ellsberg (1961) paradox. Like us, Ozdenoren and Peck resolve apparent time inconsistency by describing the decision problem as a game between an agent and a bookie (called "malevolent nature" by them). Just as we do, they point out that different games lead to different Nash equilibria, and hence different minimax optimal strategies for the agent. In particular, although the precise definitions differ, their game $\Gamma_1$ is similar in spirit to our $\mathcal{P}$-game, and their game $\Gamma_3$ is in the spirit of our $\mathcal{P}$-$X$-game.

We (as well as Ozdenoren and Peck, 2008) prove our results under the assumptions that the set of possible values of $X$ and $Y$ is finite, as is the set of actions. It would be of interest to extend this results to the case where these sets are infinite. The extension seems completely straightforward in the case that the set of values and the set of actions is countable, and we only consider bounded loss functions (i.e. $\sup_{y \in \mathcal{Y}, a \in \mathcal{A}} |L(y, a)| < \infty$). Indeed, we believe that our results should go through without change in this case, although we have not checked the details. However, once we allow an uncountable set of values, then some subtleties arise. For example, in the $\mathcal{P}$-$X$ game, we required nature to choose a value $x$ that was given positive probability by some $\Pr \in \mathcal{P}$. But there may not be such an $x$ if the set of possible values of $X$ is the interval $[0, 1]$; all the measures in $\mathcal{P}$ may then assign individual points probability 0.

We conclude this paper by giving an overview of the senses in which conditioning is optimal and the senses in which it is not, when uncertainty is represented by a set of distributions. We have established that conditioning the full set $\mathcal{P}$ on $X = x$ is minimax optimal in the $\mathcal{P}$-$x$-game, but not in the $\mathcal{P}$-game. The minimax-optimal decision rule in the $\mathcal{P}$-game is often an instance of $\mathcal{C}$-conditioning, a generalization of conditioning. The Monty Hall problem showed, however, that this is not always the case. On the other hand, if instead of the minimax criterion, we insist that update rules are sharply calibrated, then if $\mathcal{P}$ is convex, $\mathcal{C}$-conditioning is always the right thing to do after all. While, in general, $\mathcal{C}$ may depend on $\mathcal{P}$ (Theorem 6.10), if $\mathcal{P} = \langle \mathcal{P} \rangle$, we can take $\mathcal{C}(x) = \{x\}$, so standard conditioning is the "right" thing to do (Theorem 6.12).

There are two more senses in which conditioning is the right thing to do. First, Walley (1991) shows that, in a sense, conditioning is the only updating rule that is *coherent*, according to his notion of coherence. He justifies coherence decision theoretically, but not by using the minimax criterion. Note that the minimax criterion puts a total order on decision rules. That is, we can say that $\delta$ is at least as good as $\delta'$ if

$$\max_{\Pr \in \mathcal{P}} E_{\Pr}[L_\delta] \leq \max_{\Pr \in \mathcal{P}} E_{\Pr}[L_{\delta'}].$$

By way of contrast, Walley (1991) puts a partial preorder[11] on decision rules by taking $\delta$ to be at least as good as $\delta'$ if

$$\max_{\Pr \in \mathcal{P}} E_{\Pr}[L_\delta - L_{\delta'}] \leq 0.$$

Since both $\max_{\Pr \in \mathcal{P}} E_{\Pr}[L_\delta - L_{\delta'}]$ and $\max_{\Pr \in \mathcal{P}} E_{\Pr}[L_{\delta'} - L_\delta]$ may be positive, this is indeed a partial order. If we use this ordering to determine the optimal decision rule then, as Walley shows, conditioning is the only right thing to do.

Second, in this paper, we interpreted "conditioning" as conditioning the full given set of distributions $\mathcal{P}$. Then conditioning is not always an a priori minimax optimal strategy on the observation $X = x$. Alternatively, we could first somehow select a *single* $\Pr \in \mathcal{P}$, condition $\Pr$ on the observed $X = x$, and then take the optimal action relative to $\Pr \mid X = x$. It follows from Theorem 3.1 that the minimax-optimal decision rule $\delta^*$ in a $\mathcal{P}$-game can be understood this way. It defines the optimal response to the distribution $\Pr^*\in\Delta(\mathcal{X}\times\mathcal{Y})$ defined in Theorem 3.1(b)(ii). If $\mathcal{P}$ is convex, then $\Pr^*\in\mathcal{P}$. In this sense, the minimax-optimal decision rule can always be viewed as an instance of "conditioning," but on a single special $\Pr^*$ that depends on the loss function $L$ rather than on the full set $\mathcal{P}$.

---

11. For a partial order $\succeq$ is reflexive, transitive, and anti-symmetric, so that if $x \succeq y$ and $x \succeq y$, we must have $x = y$. A partial *preorder* is just reflexive and transitive.

It is worth noting that Grove and Halpern (1998) give an axiomatic characterization of conditioning sets of probabilities, based on axioms given by van Fraassen (1987, 1985) that characterize conditioning in the case that uncertainty is described by a single probability distribution. As Grove and Halpern point out, their axioms are not as compelling as those of van Fraassen. It would be interesting to know whether a similar axiomatization can be used to characterize the update notions that we have considered here.

## Acknowledgments

A preliminary version of this paper appears in *Uncertainty in Artificial Intelligence, Proceedings of the Eighteenth Conference*, 2008, with the title "A Game-Theoretic Analysis of Updating Sets of Probabilities". The present paper expands on the conference version in several ways. Most importantly, the section on calibration has been entirely rewritten, with a significant error corrected. We would like to thank Wouter Koolen, who pointed out an error in a previous version of Definition 5.3, and the anonymous referees for their thoughtful remarks. Peter Grünwald is also affiliated with Leiden University, Leiden, the Netherlands. He was supported by the IST Programme of the European Community, under the PASCAL Network of Excellence, IST-2002-506778. Joseph Halpern was supported in part by NSF under grants ITR-0325453, IIS-0534064, IIS-0812045, and IIS-0911036, by AFOSR under grant FA9550-05-1-0055 and FA9550-08-1-0438, and by ARO under grant W911NF-09-1-0281.

## Appendix A. Proofs

To prove Theorems 3.1 and Theorem 3.2, we need two preliminary observations. The first is a characterization of Nash equilibria. In the $\mathcal{P}$-game, a Nash equilibrium or saddle point amounts to a pair $(\pi^*,\delta^*)$ where $\pi^*$ is a distribution in $\mathcal{P}$ and $\delta^*$ is a randomized decision rule such that

$$
\begin{aligned}
E_{\pi^*}E_{\Pr}[L_{\delta^*}] &= \min_{\delta\in\mathcal{D}(\mathcal{X},\mathcal{A})}E_{\pi^*}[E_{\Pr}[L_\delta]]\\
&= \max_{\Pr\in\mathcal{P}}E_{\Pr}[L_{\delta^*}],
\end{aligned}
\tag{9}
$$

where $E_{\pi^*}[E_{\Pr}[L_\delta]]$ is just $\sum_{\Pr\in\mathcal{P},\pi^*(\Pr)>0}\pi^*(\Pr)E_{\Pr}[L_\delta]$. In the $\mathcal{P}$-$x$-game, a Nash equilibrium is a pair $(\pi^*,\delta^*)$ where $\pi^*$ is a distribution in $\mathcal{P}\mid X=x$ and $\delta^*$ is a randomized decision rule, such that (9) holds with $\mathcal{P}$ replaced by $\mathcal{P}\mid X=x$.

The second observation we need is the following special case of Theorem 3.2 from the work of Grünwald and Dawid (2004), itself an extension of Von Neumann's original minimax theorem.

**Theorem A.1:** *If $\mathcal{Y}'$ is a finite set, $\mathcal{P}'$ is a closed and convex subset of $\Delta(\mathcal{Y}')$, $\mathcal{A}'$ a closed and convex subset of $\mathbb{R}^k$ for some $k\in\mathbb{N}$, and $L':\mathcal{Y}'\times\mathcal{A}'\to\mathbb{R}$ is a bounded function such that, for each $y\in\mathcal{Y}'$, $L(y,a)$ is a continuous function of $a$, then there exists some $\Pr^*\in\mathcal{P}'$* and some $\rho^* \in \mathcal{A}'$ such that,

$$
\begin{aligned}
E_{\Pr^*}[L'(Y', \rho^*)] &= \min_{\rho \in \mathcal{A}'} E_{\Pr^*}[L'(Y', \rho)] \\
&= \max_{\Pr \in \mathcal{P}'} E_{\Pr}[L'(Y', \rho^*)].
\end{aligned}
\tag{10}
$$

With these observations, we are ready to prove Theorem 3.1:

**Theorem 3.1:** *Fix $\mathcal{X}$, $\mathcal{Y}$, $\mathcal{A}$, $L$, and $\mathcal{P} \subseteq \Delta(\mathcal{X} \times \mathcal{Y})$.*

(a) *The $\mathcal{P}$-game has a Nash equilibrium $(\pi^*, \delta^*)$, where $\pi^*$ is a distribution over $\mathcal{P}$ with finite support.*

(b) *If $(\pi^*, \delta^*)$ is a Nash equilibrium of the $\mathcal{P}$-game such that $\pi^*$ has finite support, then*

  (i) *for every distribution $\Pr' \in \mathcal{P}$ in the support of $\pi^*$, we have*

  $$
  E_{\Pr'}[L_{\delta^*}] = \max_{\Pr \in \mathcal{P}} E_{\Pr}[L_{\delta^*}];
  $$

  (ii) *if $\Pr^* = \sum_{\Pr \in \mathcal{P}, \pi^*(\Pr) > 0} \pi^*(\Pr) \Pr$ (i.e., $\Pr^*$ is the convex combination of the distributions in the support of $\pi^*$, weighted by their probability according to $\pi^*$), then*

  $$
  \begin{aligned}
  E_{\Pr^*}[L_{\delta^*}] &= \min_{\delta \in \mathcal{D}(\mathcal{X}, \mathcal{A})} E_{\Pr^*}[L_\delta] \\
  &= \max_{\Pr \in \mathcal{P}} \min_{\delta \in \mathcal{D}(\mathcal{X}, \mathcal{A})} E_{\Pr}[L_\delta] \\
  &= \min_{\delta \in \mathcal{D}(\mathcal{X}, \mathcal{A})} \max_{\Pr \in \mathcal{P}} E_{\Pr}[L_\delta] \\
  &= \max_{\Pr \in \mathcal{P}} E_{\Pr}[L_{\delta^*}].
  \end{aligned}
  $$

**Proof:** To prove part (a), we introduce a new loss function $L'$ that is essentially equivalent to $L$, but is designed so that Theorem A.1 can be applied. Let $\mathcal{Y}' = \mathcal{X} \times \mathcal{Y}$, let $\mathcal{A}' = \mathcal{D}(\mathcal{X}, \mathcal{A})$, and define the function $L' : \mathcal{Y}' \times \mathcal{A}' \to \mathbb{R}$ as

$$
L'((x, y), \delta) := L_\delta(x, y) = \sum_{a \in \mathcal{A}} \delta(x)(a) L(y, a).
$$

Obviously $L'$ is equivalent to $L$ in the sense that for all $\Pr \in \Delta(\mathcal{X} \times \mathcal{Y})$, for all $\delta \in \mathcal{D}(\mathcal{X}, \mathcal{A})$,

$$
E_{\Pr}[L_\delta] = E_{\Pr}[L'((X, Y), \delta)].
$$

If we view $\mathcal{A}' = \mathcal{D}(\mathcal{X}, \mathcal{A})$ as a convex subset of $\mathbb{R}^{|\mathcal{X}| \cdot (|\mathcal{A}| - 1)}$, then $L'((x, y), a)$ becomes a continuous function of $a \in \mathcal{A}'$. Let $\mathcal{P}'$ be the convex closure of $\mathcal{P}$. Since $\mathcal{X} \times \mathcal{Y}$ is finite, $\mathcal{P}'$ consists of all distributions $\Pr^*$ on $(\mathcal{X}, \mathcal{Y})$ of the form $c_1 \Pr_1 + \cdots + c_k \Pr_k$ for $k = |\mathcal{X} \times \mathcal{Y}|$, where $\Pr_1, \ldots, \Pr_k \in \mathcal{P}$ and $c_1, \ldots, c_k$ are nonnegative real coefficients such that $c_1 + \cdots + c_k = 1$. Applying Theorem A.1 to $L'$ and $\mathcal{P}'$, it follows that (10) holds for some $\Pr^* \in \mathcal{P}'$ and some $\delta^* \in \mathcal{A}' = \mathcal{D}(\mathcal{X}, \mathcal{A})$ (that is, the $\rho^*$ in (10) is $\delta^*$). Thus, there must be some distribution $\pi^*$ on $\mathcal{P}$ with finite support such that $\Pr^* = \sum_{\Pr \in \mathcal{P}, \pi^*(\Pr) > 0} \pi^*(\Pr) \Pr$. It is easy to see that the two equalities in (10) are literally the two equalities in (9). Thus, $(\pi^*, \delta^*)$ is a Nash equilibrium. This proves part (a).

To prove part (b)(i), suppose first that $(\pi^*, \delta^*)$ is a Nash equilibrium of the $\mathcal{P}$-game such that $\pi^*$ has finite support. Let $V = \max_{\Pr \in \mathcal{P}} E_{\Pr}[L_{\delta^*}]$. By (9), we have that

$$
\sum_{\Pr \in \mathcal{P}, \pi^*(\Pr) > 0} \pi^*(\Pr) E_{\Pr}[L_{\delta^*}] = V.
\tag{11}
$$

Trivially, for each $\Pr' \in \mathcal{P}$, we must have $E_{\Pr'}[L_{\delta^*}] \le \max_{\Pr \in \mathcal{P}} E_{\Pr}[L_{\delta^*}]$. If this inequality were strict for some $\Pr' \in \mathcal{P}$ in the support of $\pi^*$, then

$$\sum_{\Pr \in \mathcal{P}, \pi^*(\Pr) > 0} \pi^*(\Pr) E_{\Pr}[L_{\delta^*}] < V,$$

contradicting (11). This proves part (b)(i).

To prove part (b)(ii), note that straightforward arguments show that

$$
\begin{aligned}
& \max_{\Pr \in \mathcal{P}} E_{\Pr}[L_{\delta^*}] \\
\ge\ & \min_{\delta \in \mathcal{D}(\mathcal{X}, \mathcal{A})} \max_{\Pr \in \mathcal{P}} E_{\Pr}[L_\delta] \\
\ge\ & \max_{\Pr \in \mathcal{P}} \min_{\delta \in \mathcal{D}(\mathcal{X}, \mathcal{A})} E_{\Pr}[L_\delta] \\
\ge\ & \min_{\delta \in \mathcal{D}(\mathcal{X}, \mathcal{A})} E_{\Pr^*}[L_\delta].
\end{aligned}
$$

(The second inequality follows because, for all $\Pr' \in \mathcal{P}$, $\min_{\delta \in \mathcal{D}(\mathcal{X}, \mathcal{A})} \max_{\Pr \in \mathcal{P}} E_{\Pr}[L_\delta] \ge \min_{\delta \in \mathcal{D}(\mathcal{X}, \mathcal{A})} E_{\Pr'}[L_\delta]$.) Since $(\pi^*, \delta^*)$ is a Nash equilibrium, part (b)(ii) is immediate, using the equalities in (9). ∎

**Theorem 3.2:** *Fix $\mathcal{X}$, $\mathcal{Y}$, $\mathcal{A}$, $L$, $\mathcal{P} \subseteq \Delta(\mathcal{X} \times \mathcal{Y})$.*

(a) *The $\mathcal{P}$-x-game has a Nash equilibrium $(\pi^*, \delta^*(x))$, where $\pi^*$ is a distribution over $\mathcal{P} \mid X = x$ with finite support.*

(b) *If $(\pi^*, \delta^*(x))$ is a Nash equilibrium of the $\mathcal{P}$-x-game such that $\pi^*$ has finite support, then*

 (i) *for all $\Pr'$ in the support of $\pi^*$, we have*

$$E_{\Pr'}[L_{\delta^*}] = \max_{\Pr \in \mathcal{P} \mid X = x} E_{\Pr}[L_{\delta^*}];$$

 (ii) *if $\Pr^* = \sum_{\Pr \in \mathcal{P}, \pi^*(\Pr) > 0} \pi^*(\Pr) \Pr$, then*

$$
\begin{aligned}
& E_{\Pr^*}[L_{\delta^*}] \\
=\ & \min_{\delta \in \mathcal{D}(\mathcal{X}, \mathcal{A})} E_{\Pr^*}[L_\delta] \\
=\ & \max_{\Pr \in \mathcal{P} \mid X = x} \min_{\delta \in \mathcal{D}(\mathcal{X}, \mathcal{A})} E_{\Pr}[L_\delta] \\
=\ & \min_{\delta \in \mathcal{D}(\mathcal{X}, \mathcal{A})} \max_{\Pr \in \mathcal{P} \mid X = x} E_{\Pr}[L_\delta] \\
=\ & \max_{\Pr \in \mathcal{P} \mid X = x} E_{\Pr}[L_{\delta^*}].
\end{aligned}
$$

**Proof:** To prove part (a), we apply Theorem A.1, setting $L' = L$, $\mathcal{Y}' = \mathcal{Y}$, $\mathcal{A}' = \Delta(\mathcal{A})$, and $\mathcal{P}'$ to the convex closure of $\mathcal{P} \mid X = x$. Thus, (10) holds for some $\rho^* \in \mathcal{A}'$, which we denote $\delta^*(x)$. As in the proof of Theorem 3.1, there must be some distribution $\pi^*$ on $\mathcal{P} \mid X = x$ with finite support such that $\Pr^* = \sum_{\Pr \in \mathcal{P} \mid X = x, \pi^*(\Pr) > 0} \pi^*(\Pr) \Pr$. The remainder of the argument is identical to that in Theorem 3.1.

The proof of part (b) is completely analogous to the proof of part (b) of Theorem 3.1, and is thus omitted. ∎

**Theorem 4.4:** *Given a decision setting $DS = (\mathcal{X}, \mathcal{Y}, \mathcal{A}, \mathcal{P})$ such that $\mathcal{P} = \langle \mathcal{P} \rangle$, then for all decision probems $DP$ based on $DS$, there exists an a priori minimax-optimal rule that* *is also a posteriori minimax optimal. Indeed, every a posteriori minimax-optimal rule is also an a priori minimax-optimal rule. If, for all $\Pr \in \mathcal{P}$ and all $x \in \mathcal{X}$, $\Pr(X = x) > 0$, then for every decision problem based on DS, every a priori minimax-optimal rule is also a posteriori minimax optimal.*

**Proof:** Let $\mathcal{X}^+ = \{x \in \mathcal{X} : \max_{\Pr \in \mathcal{P}} \Pr(X = x) > 0\}$. Let $m_\delta$ be a random variable on $\mathcal{X}$ defined by taking $m_\delta(x) = 0$ if $x \notin \mathcal{X}^+$, and $m_\delta(x) = \max_{\Pr' \in \mathcal{P}|X=x} E_{\Pr'}[L_\delta]$ if $x \in \mathcal{X}^+$. We first show that for every $\delta \in \mathcal{D}(\mathcal{X}, \mathcal{A})$,

$$\max_{\Pr \in \mathcal{P}} E_{\Pr}[L_\delta] = \max_{\Pr \in \mathcal{P}} \sum_{x \in \mathcal{X}} \Pr{}_{\mathcal{X}}(X = x) m_\delta(x). \qquad (12)$$

Note that

$$
\begin{aligned}
E_{\Pr}[L_\delta] &= \textstyle\sum_{(x,y) \in \mathcal{X} \times \mathcal{Y}} \Pr((X, Y) = (x, y)) L_\delta(x, y) \\
&= \textstyle\sum_{\{x \in \mathcal{X} : \Pr{}_{\mathcal{X}}(x) > 0\}} \Pr{}_{\mathcal{X}}(X = x) \sum_{y \in \mathcal{Y}} \Pr(Y = x \mid X = x) L_\delta(x, y) \\
&= \textstyle\sum_{\{x \in \mathcal{X} : \Pr{}_{\mathcal{X}}(x) > 0\}} \Pr{}_{\mathcal{X}}(X = x) E_{\Pr|X=x}[L_\delta] \\
&\leq \textstyle\sum_{\{x \in \mathcal{X} : \Pr{}_{\mathcal{X}}(x) > 0\}} \Pr{}_{\mathcal{X}}(X = x) \max_{\Pr' \in \mathcal{P}|X=x} E_{\Pr'}[L_\delta] \\
&= \textstyle\sum_{\{x \in \mathcal{X} : \Pr{}_{\mathcal{X}}(x) > 0\}} \Pr{}_{\mathcal{X}}(X = x) m_\delta(x) \\
&= \textstyle\sum_{x \in \mathcal{X}} \Pr{}_{\mathcal{X}}(X = x) m_\delta(x).
\end{aligned}
$$

Taking the max over all $\Pr \in \mathcal{P}$, we get that

$$\max_{\Pr \in \mathcal{P}} E_{\Pr}[L_\delta] \leq \max_{\Pr \in \mathcal{P}} \sum_{x \in \mathcal{X}} \Pr{}_{\mathcal{X}}(X = x) m_\delta(x).$$

It remains to show the reverse inequality in (12). Since $\mathcal{P}$ is closed, there exists $\Pr^* \in \mathcal{P}$ such that

$$\max_{\Pr \in \mathcal{P}} \sum_{x \in \mathcal{X}} \Pr{}_{\mathcal{X}}(X = x) m_\delta(x) = \sum_{x \in \mathcal{X}} \Pr{}^*_{\mathcal{X}}(X = x) m_\delta(x).$$

Moreover, since $\mathcal{P} \mid X = x$ is closed, if $x \in \mathcal{X}^+$, there exists $\Pr^x \in \mathcal{P} \mid X = x$ such that $m_\delta(x) = E_{\Pr^x}[L_\delta]$. Define $\Pr^\dagger \in \Delta(\mathcal{X} \times \mathcal{Y})$ by taking

$$
\Pr{}^\dagger((X, Y) = (x, y)) = \left\{
\begin{array}{ll}
0 & \text{if } x \notin \mathcal{X}^+ \\
\Pr{}^*_{\mathcal{X}}(X = x) \Pr{}^x(Y = y) & \text{if } x \in \mathcal{X}^+.
\end{array}
\right.
$$

Clearly $\Pr^\dagger_{\mathcal{X}} = \Pr^*_{\mathcal{X}}$ and $(\Pr^\dagger \mid X = x) = (\Pr^x \mid X = x) \in \mathcal{P} \mid X = x$ if $x \in \mathcal{X}^+$. Thus, by definition, $\Pr^\dagger \in \langle \mathcal{P} \rangle$. Since, by assumption, $\langle \mathcal{P} \rangle = \mathcal{P}$, it follows that $\Pr^\dagger \in \mathcal{P}$. In addition, it easily follows that

$$
\begin{aligned}
&\max_{\Pr \in \mathcal{P}} \textstyle\sum_{x \in \mathcal{X}} \Pr{}_{\mathcal{X}}(X = x) m_\delta(x) \\
&= \textstyle\sum_{x \in \mathcal{X}} \Pr{}^\dagger_{\mathcal{X}}(X = x) m_\delta(x) \\
&= \textstyle\sum_{x \in \mathcal{X}^+} \Pr{}^\dagger_{\mathcal{X}}(X = x) \sum_{y \in \mathcal{Y}} \Pr{}^\dagger(Y = y \mid X = x) L_\delta(x, y) \\
&= E_{\Pr^\dagger}[L_\delta] \\
&\leq \max_{\Pr \in \mathcal{P}} E_{\Pr}[L_\delta].
\end{aligned}
$$

This establishes (12).

Now let $\delta^*$ be an a priori minimax decision rule. Since the $\mathcal{P}$-game has a Nash equilibrium (Theorem 3.1), such a $\delta^*$ must exist. Let $\mathcal{X}'$ be the set of all $x' \in \mathcal{X}$ for which $\delta^*$ is not minimax optimal in the $\mathcal{P}$–$x'$-game, i.e., $x' \in \mathcal{X}'$ iff $x \in \mathcal{X}^+$ and $\max_{\Pr' \in \mathcal{P} \mid X=x'} E_{\Pr'}[L_{\delta^*}] > \min_{\delta \in \mathcal{D}(\mathcal{X}, \mathcal{A})} \max_{\Pr' \in \mathcal{P} \mid X=x'} E_{\Pr'}[L_\delta]$. Define $\delta'$ to be a decision rule that agrees with $\delta^*$ on $\mathcal{X} \setminus \mathcal{X}'$ and is minimax optimal in the $\mathcal{P} \mid X = x'$ game for all $x' \in \mathcal{X}'$; that is, $\delta'(x) = \delta(x)$ for $x \notin \mathcal{X}'$ and, for $x \in \mathcal{X}'$,

$$\delta(x) \in \operatorname{argmin}_{\delta \in \mathcal{D}(\mathcal{X}, \mathcal{A})} \max_{\Pr' \in \mathcal{P} \mid X=x'} E_{\Pr'}[L_\delta].$$

By construction, $m_{\delta'}(x) \leq m_{\delta^*}(x)$ for all $x \in \mathcal{X}$ and $m_{\delta'}(x) < m_{\delta^*}(x)$ for all $x \in \mathcal{X}'$. Thus, using (12), we have

$$\begin{aligned}
&\max_{\Pr \in \mathcal{P}} E_{\Pr}[L_{\delta'}] \\
=\ &\max_{\Pr \in \mathcal{P}} \sum_{x \in \mathcal{X}} \Pr(X = x) m_{\delta'}(x) \\
\leq\ &\max_{\Pr \in \mathcal{P}} \sum_{x \in \mathcal{X}} \Pr(X = x) m_{\delta^*}(x) \\
=\ &\max_{\Pr \in \mathcal{P}} E_{\Pr}[L_{\delta^*}].
\end{aligned} \tag{13}$$

Thus, $\delta'$ is also an a priori minimax-optimal decision rule. But, by construction, $\delta'$ is also an a posteriori minimax-optimal decision rule, and it follows that there exists at least one decision rule (namely, $\delta'$) that is both a priori and a posteriori minimax optimal. This proves the first part of the theorem. To prove the last part, note that if $\Pr(X = x) > 0$ for all $\Pr \in \mathcal{P}$ and $x \in \mathcal{X}$, and $\mathcal{X}' \neq \emptyset$, then the inequality in (13) is strict. It follows that $\mathcal{X}'$ is empty in this case, for otherwise $\delta^*$ would not be a priori minimax optimal, contradicting our assumptions. But, if $\mathcal{X}'$ is empty, then $\delta^*$ must also be a posteriori minimax optimal.

It remains to show that every a posteriori minimax-optimal rule is also a priori minimax optimal. For all $x \in \mathcal{X}$, define $\operatorname{MM}(x) = 0$ if $x \notin \mathcal{X}^+$, and $\operatorname{MM}(x) = \min_{\delta \in \Delta} m_\delta(x)$ if $x \in \mathcal{X}^+$. Let $\Delta^*$ be the set of all a posteriori minimax-optimal rules. We have already shown that $\Delta^*$ has at least one element, say $\delta_0$, that is also a priori minimax optimal. For all $\delta \in \Delta^*$ and all $x \in \mathcal{X}$, we must have $m_\delta(x) = \operatorname{MM}(x)$. By (12), it follows that for every $\delta \in \Delta^*$,

$$\begin{aligned}
\max_{\Pr \in \mathcal{P}} E_{\Pr}[L_\delta] &= \max_{\Pr \in \mathcal{P}} \sum_{x \in \mathcal{X}} \Pr_{\mathcal{X}}(X = x) m_\delta(x) \\
&= \max_{\Pr \in \mathcal{P}} \sum_{x \in \mathcal{X}} \Pr_{\mathcal{X}}(X = x) \operatorname{MM}(x).
\end{aligned}$$

Hence,

$$\max_{\Pr \in \mathcal{P}} E_{\Pr}[L_\delta] = \max_{\Pr \in \mathcal{P}} E_{\Pr}[L_{\delta_0}].$$

Since $\delta_0$ is a priori minimax optimal, this implies that all $\delta \in \Delta^*$ are a priori minimax optimal. ∎

**Proposition 4.7:**

(a) *Every dynamically consistent decision problem is also weakly time consistent.*

(b) *Not every dynamically consistent decision problem is time consistent.*

(c) *Every strongly dynamically consistent decision problem is time consistent.*

(d) *There exist weakly time consistent decision problems that are not dynamically consistent.*

(e) *All decision problems based on $\mathcal{P}$ are dynamically consistent if and only if all decision problems based on $\mathcal{P}$ are weakly time consistent.*

**Proof:** Part (a) is immediate by part 1 of the definition of dynamic consistency. Part (b) follows because the decision problem of Example 4.5 is dynamically consistent but not time consistent. We already showed that it is not time consistent. To see that it is dynamically consistent, note that every decision rule that can be defined on the domain in the example is a priori minimax optimal, so part 1 of the definition of dynamic consistency holds automatically. Part 2 also holds automatically, since for every two decision rules $\delta$ and $\delta'$, (2) does not hold with strict inequality for $X = 0$.

For part (c), consider an arbitrary decision problem $DP$ that is strongly dynamically consistent. It is easy to construct an a posteriori minimax optimal decision rule; call it $\delta$. Since $DP$ is strongly dynamically consistent, $\delta$ must be a priori minimax optimal. Suppose, by way of contradiction, that some decision rule $\delta'$ is a priori minimax optimal but not a posteriori minimax. Since $\delta$ is a posteriori minimax optimal, it must be the case that (2) holds, and that the the inequality is strict for some $x$ with $\Pr(X = x) > 0$ for some $\Pr \in \mathcal{P}$. Thus, by strong dynamic consistency, $\delta$ must be a priori preferred to $\delta'$ according the minimax criterion, a contradiction to the assumption that $\delta'$ is a priori minimax optimal.

For part (d), consider Example 2.1 again, in which there was both time and dynamic inconsistency. Randomizing with equal probability between 0 and 1, no matter what is observed, is a posteriori preferred over all other randomized actions, but it was not the a priori minimax optimal. Now we extend the example by adding an additional action 2 and defining $L(0, 2) = L(1, 2) = -1$; $L(y, a)$ remains unchanged for $y \in \mathcal{Y}$ and $a \in \{0, 1\}$. Now both the a priori and the a posteriori minimax optimal act is to play 2, no matter what value of $X$ is observed, so time consistency holds. Yet dynamic consistency still does not hold, because after observing both $X = 0$ and $X = 1$, randomizing with equal probability between 0 and 1 is preferred over playing action 1, but before observing $X$, the decision rule that plays action 1 no matter what is observed is strictly preferred over randomizing between 0 and 1.

The "only if" direction of part (e) already follows from part (a). For the "if" direction, suppose, by way of contradiction, that all decision problems based on $\mathcal{P}$ are weakly time consistent, but some decision problem based on $\mathcal{P}$ is not dynamically consistent. This decision problem has some loss function $L$, set $\mathcal{A}$ of actions, and two decision rules $\delta$ and $\delta'$ such that $\delta$ is preferred a posteriori over $\delta'$ but not a priori; thus, in the definition of dynamic consistency, (2) holds and (3) does not. Let $L_{max}$ be the a posteriori minimax expected loss of $\delta$. Extend $\mathcal{A}$ and $L$ with an additional act $a_0$ such that for all $y$, $L(y, a_0) = L_{\max}$. Now we have a new decision problem with action set $\mathcal{A} \cup \{a_0\}$ in which $\delta$ has become a minimax optimal a posteriori rule (it is not the only one, but that does not matter). However, $\delta$ cannot be a priori minimax optimal, because (3) still does not hold for $\delta$ and $\delta'$: $\delta'$ is a priori strictly better than $\delta$. Hence, we do not have weak time consistency in this new decision problem. Since it is still a decision problems based on $\mathcal{P}$, we do not have weak time consistency for all decision problems based on $\mathcal{P}$, and we have arrived at the desired contradiction. ∎

**Theorem 5.1:** *Fix $\mathcal{X}$, $\mathcal{Y}$, $L$, $\mathcal{A}$, and $\mathcal{P} \subseteq \Delta(\mathcal{X} \times \mathcal{Y})$. If, for all $\Pr_{\mathcal{Y}} \in \mathcal{P}_{\mathcal{Y}}$, $\mathcal{P}$ contains a distribution $\Pr'$ such that $X$ and $Y$ are independent under $\Pr'$, and $\Pr'_{\mathcal{Y}} = \Pr_{\mathcal{Y}}$, then there is an a priori minimax-optimal decision rule that ignores information. Under these conditions, if $\delta$ is an a priori minimax-optimal decision rule that ignores information,* then $\delta$ essentially optimizes with respect to the marginal on $Y$; that is, $\max_{\Pr \in \mathcal{P}} E_{\Pr}[L_\delta] = \max_{\Pr_{\mathcal{Y}} \in \mathcal{P}_{\mathcal{Y}}} E_{\Pr_{\mathcal{Y}}}[L'_\delta]$.

**Proof:** Let $\mathcal{P}'$ be the subset of $\mathcal{P}$ of distributions under which $X$ and $Y$ are independent. Let $\mathcal{D}(\mathcal{X}, \mathcal{A})'$ be the subset of $\mathcal{D}(\mathcal{X}, \mathcal{A})$ of rules that ignore information. Let $\delta^* \in \mathcal{D}(\mathcal{X}, \mathcal{A})'$ be defined as the optimal decision rule that ignores information relative to $\mathcal{P}'$, i.e.

$$\max_{\Pr \in \mathcal{P}'} E_{\Pr}[L_{\delta^*}] = \min_{\delta \in \mathcal{D}(\mathcal{X}, \mathcal{A})'} \max_{\Pr \in \mathcal{P}'} E_{\Pr}[L_\delta].$$

We have

$$
\begin{aligned}
\max_{\Pr \in \mathcal{P}} E_{\Pr}[L_{\delta^*}] &\geq \min_{\delta \in \mathcal{D}(\mathcal{X}, \mathcal{A})} \max_{\Pr \in \mathcal{P}} E_{\Pr}[L_\delta] \\
&\geq \min_{\delta \in \mathcal{D}(\mathcal{X}, \mathcal{A})} \max_{\Pr \in \mathcal{P}'} E_{\Pr}[L_\delta] \\
&= \min_{\delta \in \mathcal{D}(\mathcal{X}, \mathcal{A})'} \max_{\Pr \in \mathcal{P}'} E_{\Pr}[L_\delta] \quad \text{[see below]} \\
&= \max_{\Pr \in \mathcal{P}'} E_{\Pr}[L_{\delta^*}].
\end{aligned}
\tag{14}
$$

To see that the equality between the third and fourth line in (14) holds, note that for $\Pr \in \mathcal{P}'$, we have

$$
\begin{aligned}
E_{\Pr}[L_\delta] &= \sum_{(x,y) \in \mathcal{X} \times \mathcal{Y}} \Pr(x, y) L_\delta(x, y) \\
&= \sum_{x \in \mathcal{X}} \Pr(X = x) \sum_{y \in \mathcal{Y}} \Pr(Y = y) (\sum_{a \in A} \delta(x)(a) L(y, a))
\end{aligned}
$$

The decision rule that minimizes this expression is independent of $x$; it is the distribution $\delta^*$ over actions that minimizes

$$\sum_{y \in \mathcal{Y}} \Pr(Y = y) (\sum_{a \in A} \delta^*(a) L(y, a)).$$

This calculation also shows that, since $\delta^*$ ignores information, for $\Pr \in \mathcal{P}'$, we have that

$$\max_{\Pr \in \mathcal{P}} E_{\Pr}[L_{\delta^*}] = \max_{\Pr_{\mathcal{Y}} \in \mathcal{P}_{\mathcal{Y}}} E_{\Pr_{\mathcal{Y}}}[L'_{\delta^*}] = \max_{\Pr \in \mathcal{P}'} E_{\Pr}[L_{\delta^*}]. \tag{15}$$

This implies that the first and last line of (14) are equal to each other, and therefore also equal to the second line of (14). It follows that $\delta^*$ is a priori minimax optimal. Since every a priori minimax optimal rule that ignores information must satisfy (15), the second result follows. ∎ We next prove Theorem 6.3. We first need three preliminary results.

**Lemma A.2:** If $\mathcal{P}$ is convex and $\mathcal{X}_0 \subseteq \mathcal{X}$, then $(\mathcal{P} \mid \mathcal{X}_0)_{\mathcal{Y}}$ is convex.

**Proof:** Without loss of generality, assume that $(\mathcal{P} \mid \mathcal{X}_0)_{\mathcal{Y}}$ is nonempty. Given $\Pr'_0, \Pr'_1 \in (\mathcal{P} \mid \mathcal{X}_0)_{\mathcal{Y}}$, let $\Pr'_\beta = \beta \Pr'_1 + (1 - \beta) \Pr'_0$. We show that, for all $\beta \in [0, 1]$, $\Pr'_\beta \in (\mathcal{P} \mid \mathcal{X}_0)_{\mathcal{Y}}$. Choose $\Pr_0, \Pr_1 \in \mathcal{P}$ with $\Pr_0(\mathcal{X}_0) > 0$, $\Pr_1(\mathcal{X}_0) > 0$, $(\Pr_0 \mid \mathcal{X}_0)_{\mathcal{Y}} = \Pr'_0$, and $(\Pr_1 \mid \mathcal{X}_0)_{\mathcal{Y}} = \Pr'_1$. For $c \in [0, 1]$, let $\Pr_c = c \Pr_1 + (1 - c) \Pr_0$. Then, for all $y \in \mathcal{Y}$,

$$
\begin{aligned}
\Pr_c(Y = y \mid \mathcal{X}_0) &= \frac{\Pr_c(X \in \mathcal{X}_0, Y = y)}{\Pr_c(X \in \mathcal{X}_0)} \\
&= \frac{c \Pr_1(X \in \mathcal{X}_0) \Pr_1(Y = y | X \in \mathcal{X}_0) + (1-c) \Pr_0(X \in \mathcal{X}_0) \Pr_0(Y = y | X \in \mathcal{X}_0)}{c \Pr_1(X \in \mathcal{X}_0) + (1-c) \Pr_0(X \in \mathcal{X}_0)} \\
&= \beta_c \Pr'_1(Y = y) + (1 - \beta_c) \Pr'_0(Y = y),
\end{aligned}
\tag{16}
$$

where $\beta_c = c \Pr_1(\mathcal{X}_0)/(c \Pr_1(\mathcal{X}_0) + (1-c) \Pr_0(\mathcal{X}_0))$. Clearly, $\beta_c$ is a continuous increasing function of $c$, with $\beta_0 = 0$ and $\beta_1 = 1$. Thus, there exists $c_\beta$ such that $\beta_{c_\beta} = \beta$. Since $\beta_c$ is independent of $y$, (16) holds for all $y \in \mathcal{Y}$ (with the same choice of $\beta_c$), That is, $(\Pr_{c_\beta} \mid \mathcal{X}_0)_\mathcal{Y} = \beta \Pr'_0 + (1-\beta) \Pr'_1 - \Pr'_\beta$. Thus, $\Pr'_\beta \in (\mathcal{P} \mid \mathcal{X}_0)_\mathcal{Y}$, as desired. ∎

**Lemma A.3:** *If* $\mathcal{U} = \{\mathcal{X}_1, \ldots, \mathcal{X}_k\}$ *is a collection of nonoverlapping subsets of* $\mathcal{X}$ *(i.e., for* $1 \le i < j \le k$, $\mathcal{X}_i \cap \mathcal{X}_j = \emptyset$), $(\mathcal{P} \mid \mathcal{X}_1)_\mathcal{Y}$ *is convex,* $(\mathcal{P} \mid \mathcal{X}_1)_\mathcal{Y} = (\mathcal{P} \mid \mathcal{X}_2)_\mathcal{Y} = \ldots = (\mathcal{P} \mid \mathcal{X}_k)_\mathcal{Y}$, *and* $\mathcal{V} = \bigcup_{i=1}^{k} \mathcal{X}_i$, *then for all* $j \in \{1, \ldots, k\}$, $(\mathcal{P} \mid \mathcal{V})_\mathcal{Y} \subseteq (\mathcal{P} \mid \mathcal{X}_j)_\mathcal{Y}$.

**Proof:** The result is immediate if $(\mathcal{P} \mid \mathcal{V})$ is empty. So suppose that $\Pr \in \mathcal{P}$ and $\Pr(\mathcal{V}) > 0$. Using Bayes' Rule, we have that

$$(\Pr \mid \mathcal{V})_\mathcal{Y} = \sum_{\{i : \Pr(\mathcal{X}_i \mid \mathcal{V}) > 0\}} \Pr(\mathcal{X}_i \mid \mathcal{V})(\Pr \mid \mathcal{X}_i)_\mathcal{Y}.$$

Now $(\mathcal{P} \mid \mathcal{X}_1)_\mathcal{Y} = \ldots = (\mathcal{P} \mid \mathcal{X}_k)_\mathcal{Y}$ by assumption. Thus, for all $i$ such that $\Pr(\mathcal{X}_i \mid \mathcal{V}) > 0$, there must exist some $\Pr_i \in \mathcal{P}$ such that $(\Pr \mid \mathcal{X}_i)_\mathcal{Y} = (\Pr_i \mid \mathcal{X}_1)_\mathcal{Y}$. Thus, $(\Pr \mid \mathcal{V})_\mathcal{Y} = \sum_{\{i : \Pr(\mathcal{X}_i \mid \mathcal{V}) > 0\}} \Pr(\mathcal{X}_i \mid \mathcal{V})(\Pr_i \mid \mathcal{X}_1)_\mathcal{Y}$. Since $\mathcal{P}$ is convex by assumption, by Lemma A.2, $(\mathcal{P} \mid \mathcal{X}_1)_\mathcal{Y}$ is convex as well. Thus, we can write $(\Pr \mid \mathcal{V})_\mathcal{Y}$ as a convex combination of elements of $(\mathcal{P} \mid \mathcal{X}_1)_\mathcal{Y}$, It follows that $(\Pr \mid \mathcal{V})_\mathcal{Y} \in (\mathcal{P} \mid \mathcal{X}_1)_\mathcal{Y}$. Since $(\mathcal{P} \mid \mathcal{X}_1)_\mathcal{Y} = \ldots = (\mathcal{P} \mid \mathcal{X}_k)_\mathcal{Y}$, it follows that $(\Pr \mid \mathcal{V})_\mathcal{Y} \in (\mathcal{P} \mid \mathcal{X}_j)_\mathcal{Y}$ for all $j = 1, \ldots, k$. ∎

**Lemma A.4:** *If* $\mathcal{P} = \langle \mathcal{P} \rangle$ *and* $\mathcal{U} = \{x_1, \ldots, x_k\}$, *then* $\bigcap_{j=1}^{k} (\mathcal{P} \mid X = x_j)_\mathcal{Y} \subseteq (\mathcal{P} \mid \mathcal{U})_\mathcal{Y}$.

**Proof:** Let $Q \in \bigcap_{j=1}^{k} (\mathcal{P} \mid X = x_j)_\mathcal{Y}$. There must exist $\Pr_1, \ldots, \Pr_k \in \mathcal{P}$ such that, for $j = 1, \ldots, k$, $(\Pr_j \mid X = x_j)_\mathcal{Y} = Q$. Clearly $\Pr_1(x_1) > 0$. Since $\mathcal{P} = \langle \mathcal{P} \rangle$, there also exists $\Pr \in \mathcal{P}$ such that $\Pr_\mathcal{X} = (\Pr_1)_\mathcal{X}$ and for all $j \in \{1, \ldots, k\}$ such that $\Pr_1(x_j) > 0$, we have $(\Pr \mid X = x_j)_\mathcal{Y} = (\Pr_j \mid X = x_j)_\mathcal{Y} = Q$. It follows that $(\Pr \mid \mathcal{U})_\mathcal{Y} = Q$, so $Q \in (\mathcal{P} \mid \mathcal{U})_\mathcal{Y}$. ∎

**Theorem 6.3:**

(a) *If* $\Pi$ *is* $\mathcal{C}$-*conditioning for some partition* $\mathcal{C}$ *of* $\mathcal{X}$ *and* $\mathcal{P}$ *is convex then, for all* $x \in \mathcal{X}$, *we have that* $(\mathcal{P} \mid [x]_{\Pi, \mathcal{P}})_\mathcal{Y} \subseteq \Pi(\mathcal{P}, x)_\mathcal{Y}$.

(b) *If* $\Pi$ *is standard conditioning,* $\mathcal{P} = \langle \mathcal{P} \rangle$, *and* $x \in \mathcal{X}$, *then* $\Pi(\mathcal{P}, x)_\mathcal{Y} \subseteq (\mathcal{P} \mid [x]_{\Pi, \mathcal{P}})_\mathcal{Y}$.

**Proof:** For part (a), since $\mathcal{P}$ is convex, by Lemma A.2, $(\mathcal{P} \mid \mathcal{X}')_\mathcal{Y}$ is convex for all $\mathcal{X}' \subseteq \mathcal{X}$. Let $\mathcal{U} = \{\mathcal{C}(x') \mid x' \in [x]_{\Pi, \mathcal{P}}\}$. By the definition of $[x]_{\Pi, \mathcal{P}}$, for all $x' \in [x]_{\Pi, \mathcal{P}}$, we have

$$\Pi(\mathcal{P}, x') = \mathcal{P} \mid \mathcal{C}(x') = \mathcal{P} \mid \mathcal{C}(x) = \Pi(\mathcal{P}, x).$$

Thus, by Lemma A.3, $(\mathcal{P} \mid \mathcal{V})_\mathcal{Y} \subseteq \Pi(\mathcal{P}, x)_\mathcal{Y}$, where $\mathcal{V} = \bigcup \mathcal{U} = [x]_{\Pi, \mathcal{P}} \mathcal{C}(x')$. This proves part (a).

For part (b), since $\Pi$ is standard conditioning, we have that $(\mathcal{P} \mid X = x)_\mathcal{Y} = (\mathcal{P} \mid X = x')_\mathcal{Y}$ for all $x' \in \mathcal{U}$. By assumption, $\mathcal{P} = \langle \mathcal{P} \rangle$. Thus, it follows immediately from Lemma A.4 (taking $\mathcal{U} = [x]_{\Pi, \mathcal{P}}$) that $\Pi(\mathcal{P}, x)_\mathcal{Y} \subseteq (\mathcal{P} \mid [x]_{\Pi, \mathcal{P}})_\mathcal{Y}$, as desired. ∎

We next want to prove Theorem 6.10. We first need a definition and a preliminary result.

**Definition A.5:** An update rule $\Pi$ is *semi-calibrated* relative to $\mathcal{P}$ if $(\mathcal{P} \mid [x]_{\Pi,\mathcal{P}})_{\mathcal{Y}} \subseteq \Pi(\mathcal{P}, x)_{\mathcal{Y}}$. ∎

Note that, by Theorem 6.3, if $\mathcal{P}$ is convex, then $\mathcal{C}$-conditioning is semi-calibrated for all $\mathcal{C}$.

**Lemma A.6:** *If $\Pi$ is semi-calibrated relative to $\mathcal{P}$ and $\mathcal{C} = \{[x]_{\Pi,\mathcal{P}} \mid x \in \mathcal{X}\}$, then $\mathcal{C}$ is a partition of $\mathcal{X}$ and*

(a) *$\mathcal{C}$-conditioning is narrower than $\Pi$ relative to $\mathcal{P}$.*

(b) *If $\mathcal{C}$-conditioning is not strictly narrower than $\Pi$ relative to $\mathcal{P}$, then $\Pi$ is equivalent to $\mathcal{C}$-conditioning on $\mathcal{P}$, and is calibrated.*

**Proof:** Clearly $\mathcal{C}$ is a partition of $\mathcal{X}$. For part (a), if $\Pi'$ is $\mathcal{C}$-conditioning then, by definition, $\Pi'(\mathcal{P}, x) = \mathcal{P} \mid \mathcal{C}(x) = \mathcal{P} \mid [x]_{\Pi,\mathcal{P}}$. Since $\Pi$ is semi-calibrated, $(\mathcal{P} \mid [x]_{\Pi,\mathcal{P}})_{\mathcal{Y}} \subseteq (\Pi(\mathcal{P}, x))_{\mathcal{Y}}$. Thus, $\mathcal{C}$-conditioning is narrower than $\Pi$ relative to $\mathcal{P}$.

For part (b), if $\mathcal{C}$-conditioning (i.e., $\Pi'$) is not strictly narrower than $\Pi$ relative to $\mathcal{P}$, then we must have $(\mathcal{P}(\Pi, x))_{\mathcal{Y}} = (\mathcal{P}'(\Pi, x))_{\mathcal{Y}}$ for all $x \in \mathcal{X}$, so $(\mathcal{P} \mid [x]_{\Pi,\mathcal{P}})_{\mathcal{Y}} = \Pi(\mathcal{P}, x)_{\mathcal{Y}}$, and $\Pi$ is claibrated relative to $\mathcal{P}$. ∎

**Theorem 6.10:**

(a) *$\mathcal{C}$-conditioning is sharply calibrated relative to $\mathcal{P}$ for some partition $\mathcal{C}$.*

(b) *If $\Pi$ is sharply calibrated relative to $\mathcal{P}$, then there exists some $\mathcal{C}$ such that $\Pi$ is equivalent to $\mathcal{C}$-conditioning on $\mathcal{P}$ (i.e., $\Pi(\mathcal{P}, x) = \Pi \mid \mathcal{C}(x)$ for all $x \in \mathcal{X}$).*

**Proof:** We can place a partial order $\leq_{\mathcal{P}}$ on partitions $\mathcal{C}$ by taking $\mathcal{C}_1 \leq_{\mathcal{P}} \mathcal{C}_2$ if $\mathcal{C}_1$-conditioning is narrower than $\mathcal{C}_2$ conditioning relative to $\mathcal{P}$. Since $\mathcal{X}$ is finite, there are only finitely many possible partitions of $\mathcal{X}$. Thus, there must be some minimal elements of $\leq_{\mathcal{P}}$. We claim that each minimal element of $\leq_{\mathcal{P}}$ is sharply calibrated relative to $\mathcal{P}$. For suppose that $\mathcal{C}_0$ is minimal relative to $\leq_{\mathcal{P}}$. Because $\mathcal{P}$ is convex, $\mathcal{C}_0$-conditioning is semi-calibrated (Theorem 6.3) and we can apply Lemma A.6 with $\Pi$ as $\mathcal{C}_0$. Because $\mathcal{C}_0$ is minimal, the $\mathcal{C}$ defined in Lemma A.6 cannot be strictly narrower than $\mathcal{C}_0$. It follows by Lemma A.6(b) that $\mathcal{C}_0$-conditioning is calibrated. We now show that $\mathcal{C}_0$-conditioning is in fact sharply calibrated, by showing that there exists no calibrated update rule that is a strict narrowing of $\mathcal{C}_0$-conditioning. For suppose, by way of contradiction, that $\Pi$ is an update rule that is calibrated and that is strictly narrower than $\mathcal{C}_0$ relative to $\mathcal{P}$. Then by Lemma A.6(a) there exists a partition $\mathcal{C}$ such that $\mathcal{C}$ is narrower than $\Pi$ relative to $\mathcal{P}$. But then $\mathcal{C} <_{\mathcal{P}} \mathcal{C}_0$, contradicting the minimality of $\mathcal{C}_0$. This proves part (a).

For part (b), suppose that $\Pi$ is sharply calibrated relative to $\mathcal{P}$. By Lemma A.6(a), there must be some partition $\mathcal{C}$ such that $\mathcal{C}$-conditioning is narrower than $\Pi$, relative to $\mathcal{P}$. Let $\mathcal{C}_0$ be a minimal element of $\leq_{\mathcal{P}}$ such that $\mathcal{C}_0 \leq_{\mathcal{P}} \mathcal{C}$. Part (a) shows that $\mathcal{C}_0$-conditioning is sharply calibrated relative to $\mathcal{P}$. Since $\mathcal{C}_0$-conditioning is narrower than $\Pi$, and we $\Pi$ is sharply calibrated relative to $\mathcal{P}$, we must have that $\mathcal{C}_0$-conditioning is not strictly narrower than $\Pi$ relative to $\mathcal{P}$, and hence $\Pi$ is equivalent to $\mathcal{C}_0$-conditioning on $\mathcal{P}$. ∎

**Theorem 6.12:** *If $\mathcal{P}$ is convex and $\mathcal{P} = \langle \mathcal{P} \rangle$, then standard conditioning is sharply calibrated relative to $\mathcal{P}$.*

**Proof:** By Corollary 6.4, standard conditioning is calibrated relative to $\mathcal{P}$ under the stated assumptions on $\mathcal{P}$. To show that it is sharply calibrated, suppose that there exists some update rule $\Pi'$ that is narrower than standard conditioning, and that is sharply calibrated relative to $\mathcal{P}$. By Theorem 6.10, $\Pi'$ is equivalent to $\mathcal{C}$-conditioning for some $\mathcal{C}$ relative to $\mathcal{P}$. Thus, for all $x \in \mathcal{X}$ and all $x' \in \mathcal{C}(x)$, we have that

$$(\mathcal{P} \mid \mathcal{C}(x))_{\mathcal{Y}} \subseteq (\mathcal{P} \mid x')_{\mathcal{Y}},$$

so

$$(\mathcal{P} \mid \mathcal{C}(x))_{\mathcal{Y}} \subseteq \bigcap_{x' \in \mathcal{C}(x)} (\mathcal{P} \mid x')_{\mathcal{Y}}.$$

By Lemma A.4, it is immediate that

$$\bigcap_{x' \in \mathcal{C}(x)} (\mathcal{P} \mid x')_{\mathcal{Y}} \subseteq (\mathcal{P} \mid \mathcal{C}(x))_{\mathcal{Y}}.$$

Thus, we must have

$$\bigcap_{x' \in \mathcal{C}(x)} (\mathcal{P} \mid x')_{\mathcal{Y}} = (\mathcal{P} \mid \mathcal{C}(x))_{\mathcal{Y}}. \tag{17}$$

Now we want to show that, for all $x' \in \mathcal{C}(x)$, we have that $(\mathcal{P} \mid \mathcal{C}(x))_{\mathcal{Y}} = (\mathcal{P} \mid x')_{\mathcal{Y}}$. This will show that $\mathcal{C}$ is equivalent to conditioning, and that conditioning is sharply calibrated.

Suppose not, and that $Q \in (\mathcal{P} \mid x')_{\mathcal{Y}} - \mathcal{P} \mid \mathcal{C}(x)_{\mathcal{Y}}$ for some $x' \in \mathcal{C}(x)$. Let $Q'$ be the distribution in $(\mathcal{P} \mid \mathcal{C}(x))_{\mathcal{Y}}$ that is closest to $Q$. The fact that there is such a distribution $Q'$ follows from the fact that $\mathcal{P}$ is closed (recall that we assume that $\mathcal{P}$ is closed throughout the paper). (In fact, it follows from convexity that $Q'$ is unique, but this is not necessary for our argument.) Since $Q' \in (\mathcal{P} \mid \mathcal{C}(x))_{\mathcal{Y}}$, it follows from (17) that, for each $x'' \in \mathcal{C}(x)$, there must be some distribution $\Pr_{x''} \in \mathcal{P}$ such that $\Pr_{x''}(x'') > 0$ and $(\Pr_{x''} \mid x'')_{\mathcal{Y}} = Q$. Since $\mathcal{P}$ is convex, there is some distribution $\Pr^* \in \mathcal{P}$ such that $\Pr^*(x'') > 0$ for all $x'' \in \mathcal{C}(x)$ (indeed, $\Pr^*$ can be any convex combination of the distributions $\Pr_{x''}$ for $x'' \in \mathcal{C}$ where all the coefficients are positive). Since $\mathcal{P} = \langle \mathcal{P} \rangle$, there must exist a distribution $\Pr \in \mathcal{P}$ such that $(\Pr)_{\mathcal{X}} = (\Pr^*)_{\mathcal{X}}$ (so that $\Pr$ is positive on all elements of $\mathcal{C}$), $(\Pr \mid x'')_{\mathcal{Y}} = Q'$ for all $x'' \in \mathcal{C}(x)$ other than $x'$, and $(\Pr \mid x')_{\mathcal{Y}} = Q$. Note that $(\Pr \mid (\mathcal{C}(x) - \{x'\}))_{\mathcal{Y}} = Q'$. Thus,

$$(\Pr \mid \mathcal{C}(x))_{\mathcal{Y}} = c(\Pr \mid \mathcal{C}(x) - x')_{\mathcal{Y}} + (1-c)(\Pr \mid x')_{\mathcal{Y}} = cQ' + (1-c)Q,$$

for some $c$ such that $0 < c < 1$. Clearly $cQ' + (1-c)Q$ is closer to $Q$ than $Q'$ is. This gives the desired contradiction. ∎